\documentclass{article}
\PassOptionsToPackage{sort&compress}{natbib}
\usepackage[table]{xcolor}
\usepackage{multirow}
\usepackage[preprint]{corl_2024} % Use this for the initial submission.

\usepackage{algorithmic}
\definecolor{gray}{rgb}{0.5, 0.5, 0.5}

\usepackage{amsmath}
\usepackage{amsfonts}       % blackboard math symbols
\usepackage{microtype}      % microtypography
\usepackage{bm}
\usepackage{graphicx}
\usepackage{amsfonts}
\usepackage{multicol}
\usepackage{multirow}
\usepackage{booktabs}
\usepackage{array}
\usepackage{bm}
\usepackage{xspace}
\usepackage{float}
\usepackage{wrapfig}
\makeatother
\usepackage{makecell}
\usepackage{nicefrac}
\usepackage{xfrac}
\usepackage{caption}
\usepackage{enumitem}
\usepackage{adjustbox}
\usepackage{subfig,graphicx}
\usepackage{booktabs}
\usepackage{algorithm}
\usepackage{algorithmic}
\usepackage[subtle]{savetrees}  % [subtle,moderate,extreme]

\newcommand{\customfootnotetext}[2]{{\renewcommand{\thefootnote}{#1}\footnotetext[0]{#2}}}

\definecolor{demphcolor}{RGB}{160,160,160}

\definecolor{pmcolor}{RGB}{70,70,70}
\newcommand{\pmcolor}[1]{\textcolor{pmcolor}{#1}}
\newcommand{\pms}[2]{#1{\tiny{{\pmcolor{{$\pm$#2}}}}}}

\title{Object-Centric Dexterous Manipulation\\from Human Motion Data}

\author{Yuanpei Chen$^{1, 2}$, Chen Wang$^1$, Yaodong Yang$^2$, C. Karen Liu$^1$ \\ \\ $^{1}$Stanford University, $^{2}$Peking University}

\begin{document}
\maketitle

\vspace{0.2cm}

%===============================================================================
\begin{figure*}[!h]
    \centering
    \vspace{-30pt}
    \includegraphics[width=0.9\linewidth]{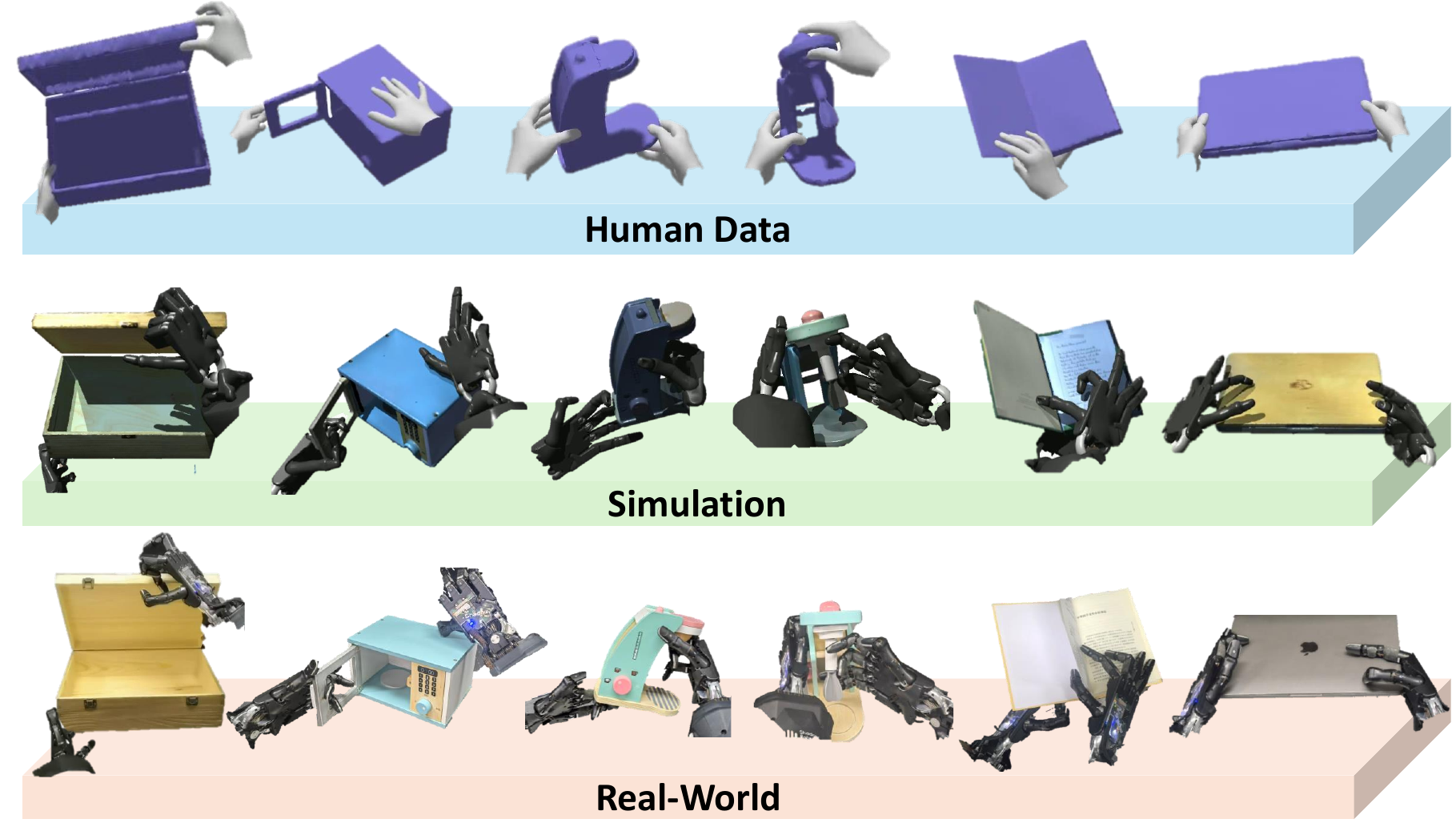}
    \caption{Our system uses human hand motion capture data and deep reinforcement learning to train dexterous robot hands for effective object-centric manipulation (i.e., learning to manipulate an object to follow a goal trajectory) in both simulation and real world.}
    \label{fig:pull}
    % \vspace{-10pt}
    \vspace{-0.4cm}
\end{figure*}

\begin{abstract} Manipulating objects to achieve desired goal states is a basic but important skill for dexterous manipulation. Human hand motions demonstrate proficient manipulation capability, providing valuable data for training robots with multi-finger hands. Despite this potential, substantial challenges arise due to the embodiment gap between human and robot hands. In this work, we introduce a hierarchical policy learning framework that uses human hand motion data for training object-centric dexterous robot manipulation. At the core of our method is a high-level trajectory generative model, learned with a large-scale human hand motion capture dataset, to synthesize human-like wrist motions conditioned on the desired object goal states. Guided by the generated wrist motions, deep reinforcement learning is further used to train a low-level finger controller that is grounded in the robot's embodiment to physically interact with the object to achieve the goal. Through extensive evaluation across 10 household objects, our approach not only demonstrates superior performance but also showcases generalization capability to novel object geometries and goal states. Furthermore, we transfer the learned policies from simulation to a real-world bimanual dexterous robot system, further demonstrating its applicability in real-world scenarios. Project website: \url{https://cypypccpy.github.io/obj-dex.github.io/}.

% Humans leverage the ability to reason about the spatial relationships between hands and objects during manipulation. Learning from human motion presents a promising approach for teaching object-centric dexterous manipulation skills to robots. However, transferring human motion data to robot actions remains challenging due to the high-dimensional action space and embodiment gap. In this paper, we explore the potential of using human motion capture data to facilitate object-centric dexterous manipulation, which is manipulating objects to follow a sequence of poses using dexterous robot hands. Our approach begins by leveraging a transformer model to generate reference trajectories for dual-handed wrist movements from human data. Subsequently, we employ reinforcement learning (RL) to synthesize nuanced dexterous manipulation policies based on the reference trajectories. The reward function is defined to minimize the distance between the current object pose and the target pose. Finally, we zero-shot transfer the trained dexterous manipulation policies to a real-world bimanual arm-hand system. Our experiments demonstrate the effectiveness of our framework in bridging the embodiment gap and its capability to generalize to unseen scenarios. Moreover, we showcase the sim-to-real transfer capability of our framework in the real robots.
\end{abstract}


\customfootnotetext{}{Correspondence to Chen Wang \textless chenwj@stanford.edu\textgreater}

%===============================================================================

\section{Introduction}
Developing bimanual multi-fingered robotic systems capable of handling complex manipulation tasks with human-level dexterity has been a longstanding goal in robotics research. When a robot engages general manipulation tasks beyond a simple pick-and-place, the definition of a \emph{task} can be difficult to establish. Previous works defined manipulation tasks in a variety of ways, including the state of the environment~\cite{tassa2018deepmind, james2020rlbench}, symbolic representations~\cite{garrett2018ffrob, garrett2021integrated}, or language descriptions~\cite{huang2022language, brohan2023can}. Regardless of how the goals are specified, a common element across these definitions is an object-centric perspective focusing on the state of the objects being manipulated. As such, the goal of our work is to train a policy for a bimanual dexterous robot to manipulate the objects according to the task goal defined as a sequence of object pose trajectories.

Prior works primarily utilize deep reinforcement learning (RL) to learn object-centric dexterous manipulation skills~\cite{huang2023dynamic, chen2022towards, zhang2023learning}. Despite the success of these methods in tasks such as in-hand object reorientation~\cite{yin2023rotating, chen2022visual, openai2019solving}, they typically focus solely on learning the movements of the fingers, neglecting the integrated coordination of both arms and hands. Training RL policy that controls both robot arms and two multi-finger hands is possible in theory, but presents substantial challenges in practice due to the high degree of freedom of the robot action space. Imitation learning (IL) can potentially tackle this challenge by leveraging the guidance from human motion data to assist policy learning. However, another challenge arises due to the morphological differences between human and robotic hands, often referred to as the ``embodiment gap''. For instance, some robotic hands are designed with only four fingers, each significantly larger than a human's, making it difficult to retarget the human hand trajectories to the robot hand while achieving the intended manipulation tasks.

One critical observation is that human finger motions are not consistently useful across various manipulation tasks due to the embodiment gap. In contrast, human wrist motions offer valuable information less sensitive to the embodiment gap, such as \textit{where} to place the palm and \textit{how} to interact with the objects in 3D space. Such motion cues significantly reduce the complexity of the high-dimensional action space in RL training, allowing it to focus on exploring finger motions to achieve the object-centric task goal. Based on this observation, we propose a hierarchical policy learning framework consisting of a high-level planner for the wrist and a low-level controller for the hand. The high-level planner is a generative-based policy, trained by imitation learning with human wrist movements, to generate robot arm actions conditioned on a desired trajectory of the object's movements. Based on the generated arm motions, the low-level controller outputs fine-grained finger actions learned through RL exploration rather than imitation of human data. In addition, the RL policy also learns the residual wrist movements to integrate with arm action into the final robot action. The reward function for the RL training is the likelihood between the object's movements during interaction and the reference trajectory. By harnessing the strengths of both RL and IL, we enable the robot to adapt to its own hand embodiment while keeping the training tractable by refining the action space using human data.

To ensure that the learned policy can adapt to a variety of reference object trajectories, not just a single scenario, we utilize ARCTIC~\cite{fan2023arctic}, a comprehensive dataset with 51 hours of diverse hand-object manipulation motion capture sequences. Our experiments demonstrate that the learned policy exhibits generalization to novel object geometries and unseen motion trajectories. In addition, we successfully transfer our policy from simulation environments to a real-world bimanual dexterous robot, further validating its practical applicability in real-world manipulation tasks.

\section{Related Works}

\subsection{Dexterous Manipulation}
Dexterous manipulation is a long-standing research topic in robotics~\cite{bai2014dexterous, kumar2014real, mason1985robot, mordatch2012contact}. Traditional methods rely on analytical dynamic models for trajectory optimization~\cite{chen2024springgrasp, mordatch2012contact, bai2014dexterous, kumar2014real}, which fall short in complex tasks due to the simplification of contact dynamics.~\cite{pang2023global} and~\cite{cheng2023enhancing} propose model-based optimization methods to solve the contact-rich dexterous manipulation task, but a correct dynamics model of the environment is required. Recently, deep reinforcement learning (RL) has showcased promising results in training dexterous manipulation skills such as in-hand object reorientation~\cite{chen2021system, chen2022visual, yin2023rotating, qi2023general, qi2023hand, dasari2023learning, openai2019solving, yang2024anyrotate, pitz2023dextrous, handa2022dextreme, khandate2023sampling}, bimanual manipulation~\cite{huang2023dynamic, lin2024twisting, chen2022towards}, sequential manipulation~\cite{xu2023dexterous, chen2023sequential, gupta2021reset}, and human-like activities~\cite{zakka2023robopianist}. \cite{dasari2023learning} used pre-grasp techniques to learn a grasping policy to make the object follows a manual designed trajectory, but we focus on more challenging tasks, such as articulation manipulation, bimanual manipulation. Despite the progress, successfully training a dexterous RL policy often requires extensive reward engineering and system design, which limits its practicality in some scenarios. Besides RL, imitation learning (IL) is also widely used for training dexterous policies~\cite{rajeswaran2017learning, radosavovic2021state}. By performing supervised-learning with human teleoperation data~\cite{arunachalam2023holo, guzey2023dexterity, handa2020dexpilot, sivakumar2022robotic, qin2023anyteleop}, prior works show impressive results in dexterous grasping~\cite{mandikal2021learning, chen2022learning} and general manipulation tasks~\cite{qin2022dexmv, cui2022play, haldar2023teach, qin2022one, arunachalam2022dexterous, guzey2023see, lin2024learning, ze20243d}. However, teleoperation data are often expensive to collect due to the requirement of a real-world robot system. In this work, we focus on learning directly from human hand motion capture data without the need for teleoperation systems.

\subsection{Learning from Human Motion}

Recently, learning from human motion data has started to receive more attention because it allows scaling up data collection without robot hardware. Prior works leverage human videos~\cite{bahl2022human, mandikal2022dexvip, bahl2023affordances, bharadhwaj2023towards, shaw2023videodex, bahety2024screwmimic}, motion capture data~\cite{zhang2023artigrasp, chen2022dextransfer, wang2023mimicplay, garcia2018first, hasson2019learning, taheri2020grab} to extract valuable motion hints for manipulation such as trajectory-level plans ~\cite{bahl2022human, wang2023mimicplay}, object affordance~\cite{bahl2023affordances} and motion priors~\cite{mandikal2022dexvip, shaw2023videodex}. For dexterous manipulation, VideoDex~\cite{shaw2023videodex}, DexMV~\cite{qin2022dexmv}, DexTransfer~\cite{chen2022dextransfer} and DexCap~\cite{wang2024dexcap} showcase the potential of using analytical methods (e.g., inverse kinematics) to retarget human hand motion to robot hardware, such as matching joint angles~\cite{shaw2023videodex, qin2022dexmv}, fingertip positions~\cite{wang2024dexcap}. ~\cite{ye2012synthesis} showing independent control of the fingers and arm for capturing detailed hand manipulation of objects. DexPilot~\cite{handa2020dexpilot} formulate the retargeting objective as a non-linear optimization problem and retarget human motion to robot by minimizing a cost function. However, due to the embodiment gap between human and robot hands, position-based retargeting methods do not guarantee the replication of task success. In contrast, our approach uses human data as guidance for RL training, which learns the motion retargeting conditioned on the robot's embodiment. 
Notably, ~\cite{ho2016generative, rajeswaran2017learning, peng2018deepmimic, peng2021amp, peng2022ase, wang2023physhoi} share the same idea of utilizing human data as guidance or reward for reinforcement learning. While these works focus on learning human whole-body control in simulation, we study dexterous manipulation with multi-fingered robotic hands and transfer the learned policy from simulation to the real world.

\section{Task Formulation}

\begin{figure*}[!t]
    \centering
    \includegraphics[width=1.0\linewidth]{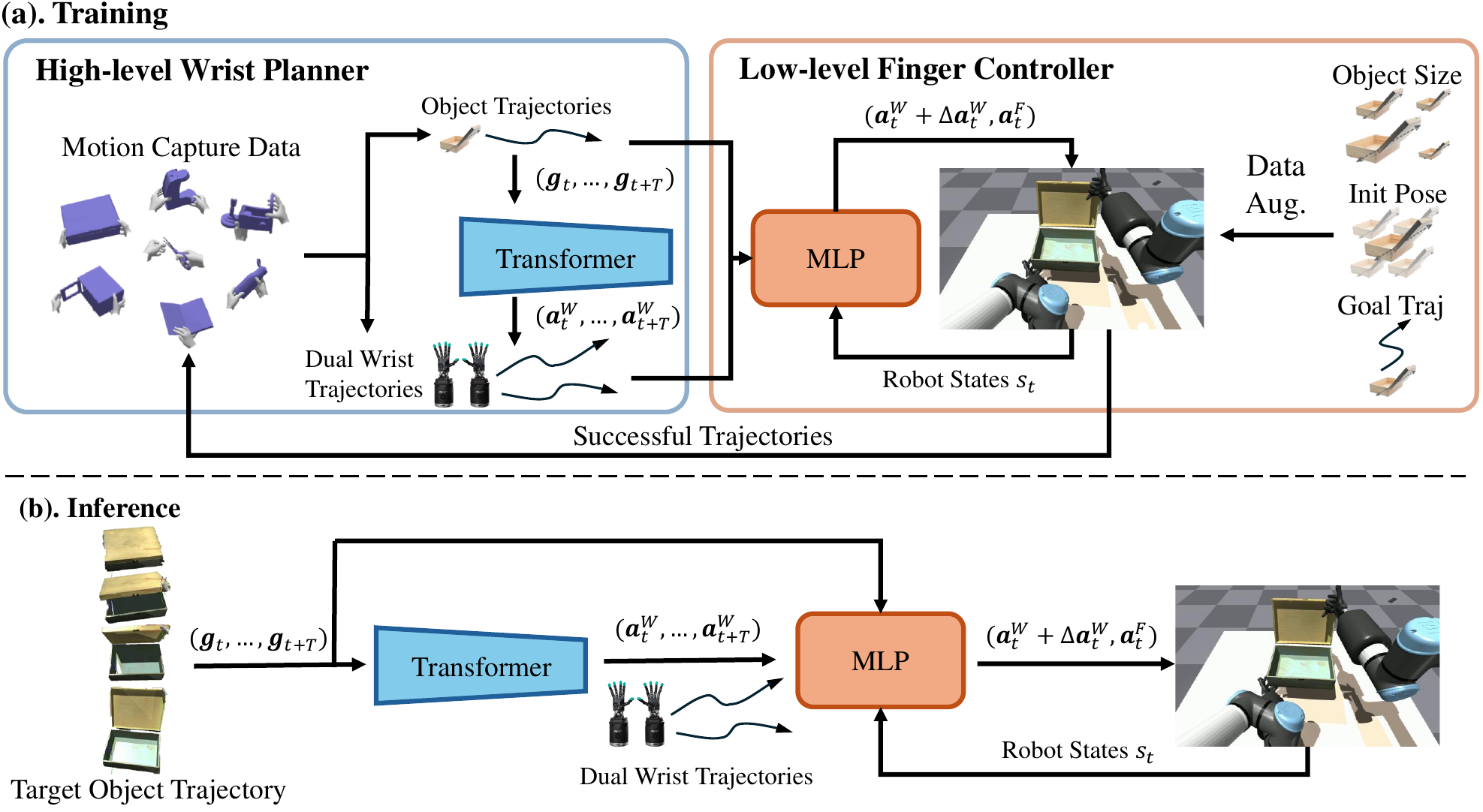}
    \caption{Overview of our framework. (A) Training: Firstly, we use human motion capture data to train a generation model to synthesize dual hand trajectory conditions on object trajectory. Then we use the RL to train a low-level robot controller conditioned on the dual hand trajectory generated by the trained high-level planner. During this process we augment the data in simulation to improve the high-level planner and low-level controller simultaneously (B) Inference: Given a single object goal trajectory, our framework generates dual hand reference trajectory and guides the low-level controller to accomplish the task. }

    \label{fig:method}
    \vspace{-15pt}
\end{figure*}

The goal of an object-centric manipulation task is to let the robot physically interact with the object to achieve the desired motion trajectory. We define the motion trajectory as the sequence of the object's $SE(3)$ transformation $G=(\bm{g}_1,\bm{g}_2, \dots ,\bm{g}_T)$, 

where each time step $\bm{g}_i = (\bm{g}_i^R, \bm{g}_i^T, g^J_i)$ consists a 3D rotation $\bm{g}_i^R$, a 3D translation $\bm{g}_i^T$, and the joint angle $g_i^J$. $g^J_i$ can be omitted if the object is a single rigid body.

We then formulate an object-centric manipulation task as a Markov Decision Process (MDP) $\mathcal{M} = (\pmb{S}, \pmb{A}, \pi, \mathcal{T}, R, \gamma, \rho, G)$, where $\pmb{S}$ is the state space, $\pmb{A}$ is the action space, $\pi$ is the agent's policy, $\mathcal{T}(\bm{s}_{t+1}|\bm{s}_t, \bm{a}_t)$ is the transition distribution, $R$ is the reward function, $\gamma$ is the discount factor, and $\rho$ is the initial state distribution. The policy $\pi$ conditions on the reference object state trajectory $G$ and the current state $\bm{s}_t$, and generates robot action distributions $\bm{a}_t$ to maximize the likelihood between the future object states $(\bm{s}_{t+1},\bm{s}_{t+2}, \dots ,\bm{s}_{t+T})$ and the reference trajectory $G$. This formulation is versatile to adapt to downstream tasks, such as moving objects to desired locations and in-hand object re-orientation with bi-manual hands. 

\section{Method}
In this section, we introduce our framework for object-centric manipulation. The overview of the framework is shown in Figure~\ref{fig:method}. Our framework consists of three parts: high-level planner (Section~\ref{sec:high-level}), low-level controller (Section~\ref{sec:low-level}) and the data augmentation loop (Section~\ref{sec:plil}). The details of our sim-to-real policy transfer are introduced in Section~\ref{sec:sim2real}.

\subsection{High-Level Planner}
\label{sec:high-level}

One of the key challenges in training a dexterous robot policy, especially with bi-manual dexterous hands, is managing the high-dimensional action space. Despite the embodiment gap between human and robot hands, human hand motion, particularly wrist movements, provides highly informative hints for how to interact with objects and environments. Such motion hints not only narrow the action space for the robot but also provide guidance toward the task goal. Based on this observation, we first train a generative model to synthesize wrist motion by imitating human wrist movements in the motion capture data.

We train a Transformer-based generative model $\pi^H$ that takes object category ID $c$, and the desired object motion trajectory $G = (\bm{g}_t,\bm{g}_{t+1},..., \bm{g}_{t+T})$ as inputs and outputs a sequence of 6-DoF wrist actions $(\bm{a}^W_t,\bm{a}^W_{t+1},..., \bm{a}^W_{t+T})$, where each action $\bm{a}^W_i = (\bm{p}_i^{l}, \bm{p}_i^{r})$ consists the 6-DoF pose of the left hand $\bm{p}_i^{l}$ and right hand $\bm{p}_i^{r}$ in SE(3). In our experiments, we use $T = 10$. We leverage the entire ARCTIC dataset~\cite{fan2023arctic} to train $\pi^H$ with a behavior cloning algorithm. The total training samples consist of 339 sequences of human hand mocap trajectories, totaling 2.1 million steps.

\subsection{Low-Level Controller}
\label{sec:low-level}

Based on the generated wrist trajectory from the high-level planner $\pi^H$, the low-level controller $\pi^L$ can start from a reasonable wrist pose and focus on learning the fine-grained finger motions to physically interact with the object to achieve the task goal $G$. We use Proximal Policy Optimization (PPO)~\cite{schulman2017proximal} to train $\pi^L$. The policy $\pi^L$ takes the current observation $\bm{s}_i$ 
% \karen{Exactly what $\bm{s}_i$ is has not been mentioned.}
, the desired object motion trajectory $G = (\bm{g}_t,\bm{g}_{t+1},..., \bm{g}_{t+T})$, and a sequence of 6-DoF wrist actions $(\bm{a}^W_t,\bm{a}^W_{t+1},..., \bm{a}^W_{t+T})$ generated by high-level planner as inputs, and outputs the finger joint action $\bm{a}^F_t$. Here the observation $\bm{s}_t$ contains the object pose and robot proprioception. The reward function is defined as $r_t=\exp^{ -(\lambda_1*\Vert \bm{g}^R_t - \bm{\hat{g}}_t^R \Vert_2+\lambda_2*\Vert \bm{g}^T_t - \bm{\hat{g}}_t^T \Vert_2+\lambda_3*\Vert g^J_t - \hat{g}_t^J \Vert_2)}$, aiming to minimize the distance between object's movements and the desired goal trajectory. $\bm{\hat{g}}_t^R$, $\bm{\hat{g}}_t^T$ and $\hat{g}_t^J$ is the current 3D translation, 3D rotation and joint angle of the object respectively. $\lambda_1$, $\lambda_2$, and $\lambda_3$ are the hyperparameters to balance the weight of each component of the reward. In some scenarios, when the robot's hand is larger than the human's, the generated wrist actions from the high-level planner $\pi^H$ need to be adjusted accordingly. To achieve this, $\pi^L$ learns to output a residual wrist action $\Delta \bm{a}^W_t$ within a fixed range ($\pm 4$ centimeters for transition and $\pm 0.5$ radian for rotation). With the residual wrist actions, the policy can now adjust the wrist position to better synthesize the motion of the robot hand. The final robot action is a combination of $(\bm{a}^W_t + \Delta \bm{a}^W_t, \bm{a}^F_t)$. Please refer to Appendix~\ref{app:RL_implement} for more detail about the observation space and the reward function.

\subsection{Data Augmentation for Generalization}
\label{sec:plil}

Training policy only with the ARCTIC dataset~\cite{fan2023arctic} limits the diversity of task objects and generated motions. For instance, given a box twice the size of the one used in the dataset, it is challenging for the learned policies to manipulate the box effectively because of the unseen object geometry. To improve the generalization capability, it is critical to augment the data and train the policies to generalize in these scenarios. Thereby, we propose Data Augmentation Loop (DAL) to handle different object geometries and goal trajectories during training $\pi^L$.

Specifically, we introduce three types of augmentation during the RL training of the $\pi^L$: randomizing the object's size in all three dimensions (width, length, height), randomizing the object's initial pose, and modifying the goal trajectories of the object with waypoint interpolation. The low-level controller then learns to adapt to the augmented scenarios during the RL training. Finally, we collect the successful motion trajectories executed by the low-level controller and add them to the training dataset to fine-tune the high-level planner. By leveraging the adaptability of the RL training process, we can synthesize novel motion trajectories to improve the policy’s generalization beyond the scope of the original training data. 
Detailed implementations of the Data Augmentation Loop can be found in Appendix~\ref{app:dal}.

\subsection{Sim-to-Real Transfer}
\label{sec:sim2real}

When deploying the policy to the real world, some observation cannot be accurately estimated, such as joint velocity and object velocity. We use the teacher-student policy distillation framework~\cite{chen2022visual, chen2021system, rusu2015policy} to remove the dependence on these observation inputs from the policy. In real-world deployments, our system uses four top-down cameras to perform articulated object pose estimation. We use FoundationPose~\cite{wen2023foundationpose} to estimate the 6-DoF pose of different parts of the articulated object and calculate the joint angle between them. The entire object pose tracking system runs at the speed of 15Hz. The learned policies generate actions to control the robot, and the low-level controller is processed by a low-pass filter with an exponential moving average (EMA) smoothing factor~\cite{chen2022visual} to reduce the jittering motions of the robot fingers.

For more details on the sim-to-real setups, please refer to Appendix~\ref{app:real_implement}.

\section{Experiments}

\begin{figure*}[!t]
    \centering
    \includegraphics[width=1.0\linewidth]{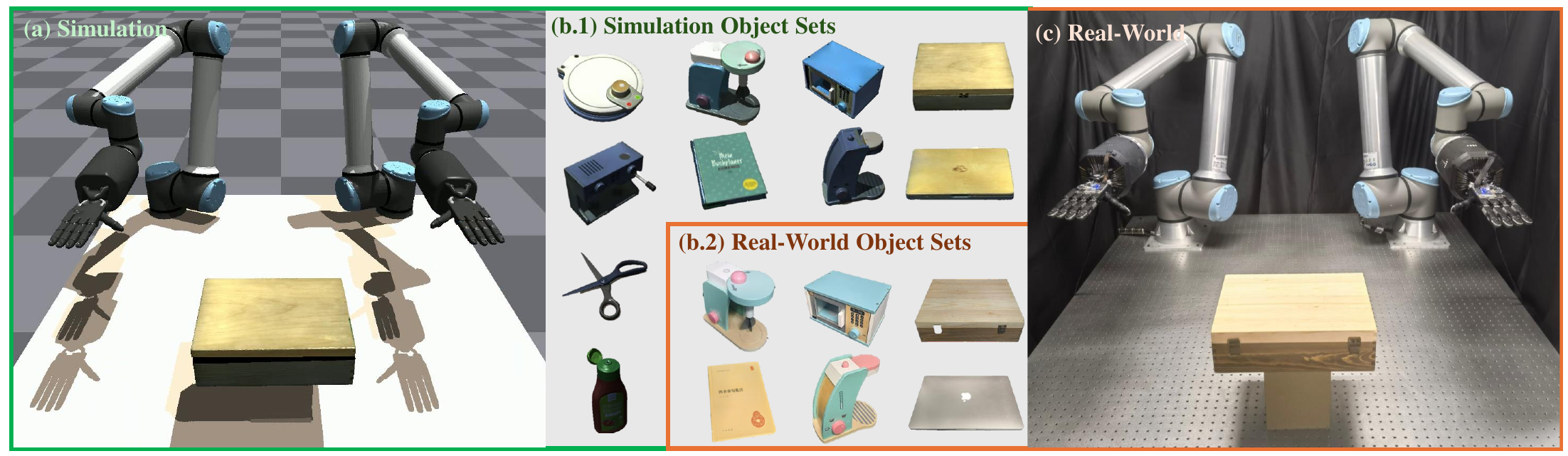}
    \caption{Overview of the environment setups. (a) Workspace of the simulation. We employ two Shadow Hands, each individually mounted on separate UR10e robots, arranged in an abreast configuration. (b.1) Object sets in the simulation. (b.2) Object sets in the real-world.  (c) Workspace of the real-world, mirroring the simulation, the robot system uses the same Shadow Hands and UR10e robots as the simulation.}
    \label{fig:workspace}
    \vspace{-8pt}
    % \vspace{-0.4cm}
\end{figure*}

The experiments are designed to answer the following research questions: (1) Can the high-level planner generalize to unseen trajectories and unseen objects? 
(Sec.~\ref{sec:high-level_results}) (2) Does our hierarchical approach help bridge the embodiment gap between human and robot hands? (Sec.~\ref{sec:effectiveness_results}) (3) Can our trained policy generalize to unseen object geometries and goal trajectories? (Sec.~\ref{sec:generalization_results}) (4) Can we transfer the policy from simulation to a real-world bimanual dexterous robot system? (Sec.~\ref{sec:sim2real_results}). In this section, we first introduce our training data and baseline setups, followed by the results for each research question. The experimental setups are shown in Figure.\ref{fig:workspace}. Our simulation experiments are all evaluated in 10 different seeds, and the real-world experiments are all evaluated in 20 different trials. The small numbers in each table represent the standard deviations across different seeds and trials.

\begin{figure}

\begin{minipage}[!t]{0.48\textwidth}

\centering
\begin{tabular}{c|c|ccc}
\toprule
                             &         & MLP     & RNN & Ours  \\ \midrule
\multirow{2}{*}{Trained}     & TE & \pms{5.4}{0.2}   &   \pms{5.0}{0.1}     & \pms{\textbf{4.2}}{0.2}       \\
                             & OE & \pms{14.6}{0.7}  &    \pms{12.4}{0.8}    & \pms{\textbf{9.6}}{0.6}        \\ \midrule
Unseen & TE & \pms{5.6}{0.5}   & \pms{6.2}{0.3}      & \pms{\textbf{5.0}}{0.8}      \\
Traj                             & OE & \pms{9.4}{1.0}   &   \pms{9.2}{0.5}     & \pms{\textbf{7.2}}{0.7}      \\ \midrule
Unseen  & TE & \pms{18.2}{1.5}  &    \pms{17.7}{1.2}    & \pms{\textbf{12.4}}{1.1}      \\
Object                             & OE & \pms{109.4}{5.8} &     \pms{82.2}{4.2}   & \pms{\textbf{75.5}}{4.7}      \\
\bottomrule

\end{tabular}
% \vspace{5pt}
\captionof{table}{Results for the high-level planner}
\label{tab:table_1}

\end{minipage}
\hspace{0.2cm}
\begin{minipage}[!t]{0.48\textwidth}

\centering
\label{tab:my-table}
\begin{tabular}{l|cccc}
\toprule
& \multirow{2}{*}{\shortstack{Fingertip \\ Mapping}} & \multirow{2}{*}{\shortstack{Vanilla \\RL}} & \multirow{2}{*}{Ours} \\
 &   
  \\
\midrule
Box               & \pms{13.1}{1.8}         & \pms{20.4}{1.8}       &  \pms{\textbf{69.8}}{6.6}  \\
Micro.         &   \pms{60.6}{2.7}       &   \pms{56.1}{9.8}     &  \pms{\textbf{100}}{0.0} \\
Laptop            & \pms{9.2}{0.5}         & \pms{8.6}{1.4}       & \pms{\textbf{76.7}}{4.1}   \\
Coffee.    & \pms{8.2}{0.6}         & \pms{9.2}{3.1}       & \pms{\textbf{74.8}}{3.8}   \\
Mixer             & \pms{8.3}{2.1}         &  \pms{10.8}{3.2}      & \pms{\textbf{82.8}}{2.1}   \\
Notebook          &    \pms{4.5}{0.1}      &    \pms{4.5}{0.3}    &   \pms{\textbf{64.3}}{8.4} \\
\bottomrule

\end{tabular}
% \vspace{5pt}
\captionof{table}{Results for the real-world experiments}
\label{tab:table_2}

\end{minipage}
\vspace{-0.3cm}

\end{figure}

\begin{table}[t]
\centering
% \small
\begin{tabular}{l|cccccc}
\toprule
 % &        & Finger & Joint & Vanilla & Ours        & Ours & Ours \\
 %  &        & IK & IK & RL & (w. FR)       & (w.o. PCIL)    &

& \multirow{2}{*}{\shortstack{Fingertip \\ Mapping}} & \multirow{2}{*}{\shortstack{Finger Joint \\ Mapping}} & \multirow{2}{*}{\shortstack{Vanilla \\RL}} & \multirow{2}{*}{\shortstack{Ours \\ (w. FR)}} & \multirow{2}{*}{\shortstack{Ours \\ (w.o. DAL)}} & \multirow{2}{*}{Ours} \\
 &   
  \\ \midrule
Box & \pms{14.6}{0.3} & \pms{8.9}{0.2} & \pms{23.5}{3.5} & \pms{56.2}{7.4} &  \pms{100}{0.0} & \pms{\textbf{100}}{0.0} \\

Coffee Maker & \pms{9.3}{0.6}    & \pms{9.0}{0.5}    & \pms{10.7}{2.7}    & \pms{78.6}{1.6}               & \pms{71.5}{2.6} & \pms{\textbf{86.1}}{5.5} \\

Espresso Machine& \pms{22.2}{0.4}      & \pms{7.0}{0.8}  & \pms{14.3}{1.5} & \pms{70.7}{3.5} &  \pms{75.4}{4.3}       & \pms{\textbf{81.1}}{8.6} \\

Ketchup& \pms{14.8}{0.7}   & \pms{9.5}{0.2}    & \pms{4.9}{2.7}    & \pms{15.2}{1.7}       & \pms{21.8}{7.2}       & \pms{\textbf{41.2}}{13.3} \\

Microwave & \pms{38.7}{0.2} & \pms{27.5}{0.6} & \pms{43.5}{2.4} & \pms{61.2}{5.3} &  \pms{100}{0.0} & \pms{\textbf{100}}{0.0} \\

Mixer & \pms{21.7}{0.9}    & \pms{10.7}{0.8}    & \pms{42.1}{1.4}    & \pms{42.2}{4.0}               & \pms{44.2}{6.4} & \pms{\textbf{57.6}}{4.9} \\

Notebook& \pms{10.1}{0.5}       & \pms{5.9}{0.4}  & \pms{10.6}{2.9} & \pms{31.1}{4.1} &  \pms{38.1}{4.8}       & \pms{\textbf{38.7}}{3.3} \\

Scissors & \pms{4.2}{0.5}   & \pms{4.1}{0.6}    & \pms{4.4}{0.6}    & \pms{20.7}{2.0}       & \pms{35.9}{4.0}       & \pms{\textbf{41.4}}{14.9} \\

Laptop & \pms{9.9}{0.4}   & \pms{8.8}{1.1}    & \pms{33.0}{2.1}    & \pms{42.5}{5.4}       & \pms{100}{0.0}       & \pms{\textbf{100}}{0.0} \\

\bottomrule
\end{tabular}
\vspace{5pt}
\caption{Results for the experiments of using one policy per object.}
\label{tab:table_3}
\vspace{-0.5cm}
\end{table}

\textbf{Data.} The ARCTIC dataset consists of 10 objects, each with 20 sequences of motion capture data. We choose 9 objects, each with 16 sequences of data, as our training set. We left the remaining objects and motion sequences as the unseen testing set.

\textbf{Baselines.} We compare our approach with the following methods and ablations: (1) \textit{Finger Joint Mapping}: Match the finger joint angles between human and robot hands as demonstration and perform end-to-end imitation learning~\cite{qin2022dexmv}.
(2) \textit{Fingertip Mapping}: Match the fingertip positions between human and robot hands using IK as demonstration and perform end-to-end imitation learning~\cite{wang2024dexcap}. (3) \textit{Vanilla RL}: Train a PPO policy~\cite{schulman2017proximal} to learn the motions of the arm and hand together. (4) \textit{Ours}: Full implementation of our hierarchical policy learning approach. (5) \textit{Ours w. FR}: Our method with an additional fingertip-matching reward during training.
(6). \textit{Ours w/o. DAL}: Our method without data augmentation loop. 

\subsection{Performance of the high-level planner}
\label{sec:high-level_results}

\textbf{Tasks.} To validate the accuracy of the learned high-level planner, we introduce three experimental tasks: (1) \textit{Trained}: Testing the policy with trained objects and goal trajectories. (2) \textit{Unseen Traj}: Testing the policy conditioned on unseen goal trajectories with trained objects. (3) \textit{Unseen Obj}: Testing the policy with unseen objects and trained goal trajectories.

\textbf{Metric.} 
Evaluation metric is defined as the distance between the ground truth wrist pose trajectory of the hand and the wrist pose trajectory of the left and right hand output by our high-level planner.
For each step, we compute translation error (TE) using 
Euclidean distance in centimeters and orientation error (OE) 
using angle difference.
We report the cumulative error of the entire motion sequence in Table~\ref{tab:table_1}

\textbf{Results.} Table~\ref{tab:table_1} shows that Ours performs the best in generating wrist motions, with the lowest cumulative translation and orientation error. More importantly, the performance does not drop when testing with unseen goal trajectories (\textit{Unseen Traj}), which demonstrates the generalization capability of our high-level planner to novel object-centric task goals. 

\subsection{Effectiveness of learning from human with hierarchical pipeline}
\label{sec:effectiveness_results}

\textbf{Task.} To validate the effectiveness of our framework, we first train one policy conditioned on a single trajectory in the training set for each object and test its rollout performance. Each policy is trained on a single object.

\textbf{Metric.} We use \textbf{Completion Rate} as the evaluation metric, which indicates the percentage of the goal object trajectory completed by the policy. Completion is achieved if and only if the object's pose error is smaller than a threshold ($5$ centimeters in translation, $2.5$ centimeters in object's longest dimension multiplied by rotation angle, and $0.5$ radians in joint angle).
Each goal trajectory has a total of 500 action steps.

\begin{wrapfigure}[9]{r}{6.9cm}
    \centering
    \vspace{-1.1cm}
    \setlength{\abovecaptionskip}{0.1cm}
    \includegraphics[width=1.0\linewidth]{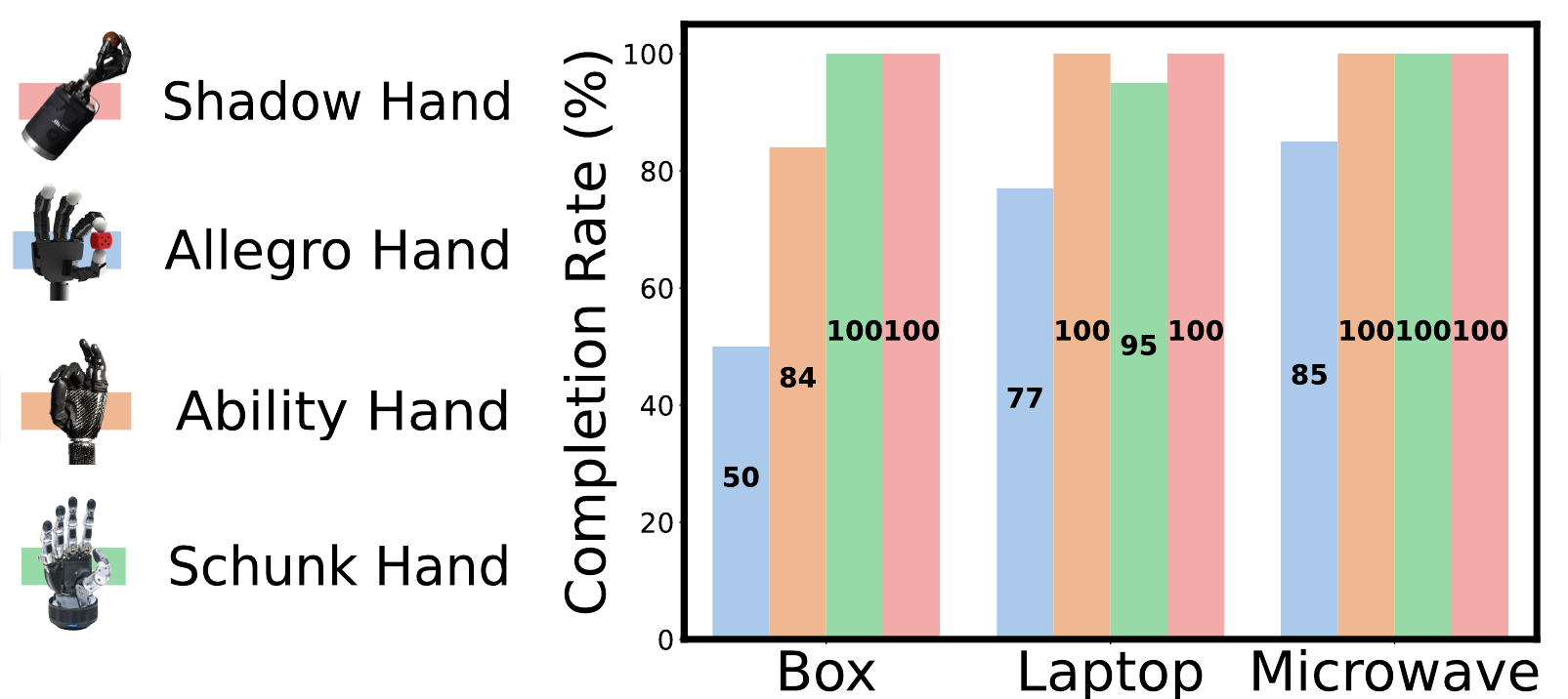}
    \caption{Experiments on the different embodiment. We study four types of the dexterous hand in three tasks from the Section~\ref{sec:effectiveness_results}.}
    
\label{fig:replan}
\end{wrapfigure}

\textbf{Results.} Table~\ref{tab:table_3} demonstrates that our hierarchical learning framework outperforms traditional hand pose matching methods (\textit{Finger Joint Mapping}, \textit{Fingertip Mapping}) by 50\% in completion rate, indicating that our low-level RL significantly helps in bridging the embodiment gap when learning from human data. Moreover, \textit{Ours} surpasses \textit{Vanilla RL} by 47.3\% on average, underscoring the challenge of training arm and hand actions together with RL, and emphasizing the advantage of our high-level planner for guiding the RL in high-dimensional action space. Additionally, the inclusion of human finger motions in the RL reward (\textit{Ours (w. FR)}) does not yield benefits and even leads to lower performance, validating our hypothesis that the embodiment gap makes human finger motion unsuitable for training robot actions. Lastly, our data augmentation loop further brings an additional 7\% improvement (\textit{Ours} vs. \textit{Ours (w.o. DAL)}). We further apply our method to four different types of multi-fingered dexterous hands, varying in size and degree of freedom. The results are shown in Figure~\ref{fig:replan}. Our method achieved more than 50\% completion rate for all hands, demonstrating that our framework can effectively transfer human data to different robot hand embodiments.

\subsection{Generalization to unseen scenarios}
\label{sec:generalization_results}

\textbf{Tasks.} We design four types of tasks to test the policy's generalization capability: (1). \textit{Single Obj - Trained Traj}: One policy for each object, and testing with trained goal trajectories. (2). \textit{Single Obj - Unseen Traj}: Same as prior but testing the policy conditioned on unseen goal trajectories. (3). \textit{Multi Obj - Trained Obj}: One policy trained with all objects, and testing with trained objects. (4). \textit{Multi Obj - Unseen Obj}: Same as prior but testing the policy with unseen objects. We use the completion rate as the evaluation metric same as in Section~\ref{sec:effectiveness_results}.

\textbf{Results.} In Table~\ref{tab:table_4}, our algorithm surpassing the results of \textit{Vanilla RL} on \textit{Single Obj - Unseen Traj} and \textit{Multi Obj - Unseen Obj} by more than 28\%. This indicates that our hierarchical structure substantially improves generalization capabilities across unseen trajectories and unseen object geometries. Notably, unlike the Section~\ref{sec:effectiveness_results}, we observe an average 18\% improvement in Table~\ref{tab:table_3} with our DAL (\textit{Ours} vs. \textit{Ours (w.o. DAL)}), showcasing that the DAL greatly helps generalization. Traditional mapping methods (\textit{Finger Joint Mapping}, \textit{Fingertip Mapping}) and \textit{Ours (w. FR)} cannot generalize to unseen trajectories and unseen object geometries due to their dependency on the finger information.

\begin{figure*}[!t]
    \centering
    \includegraphics[width=1.0\linewidth]{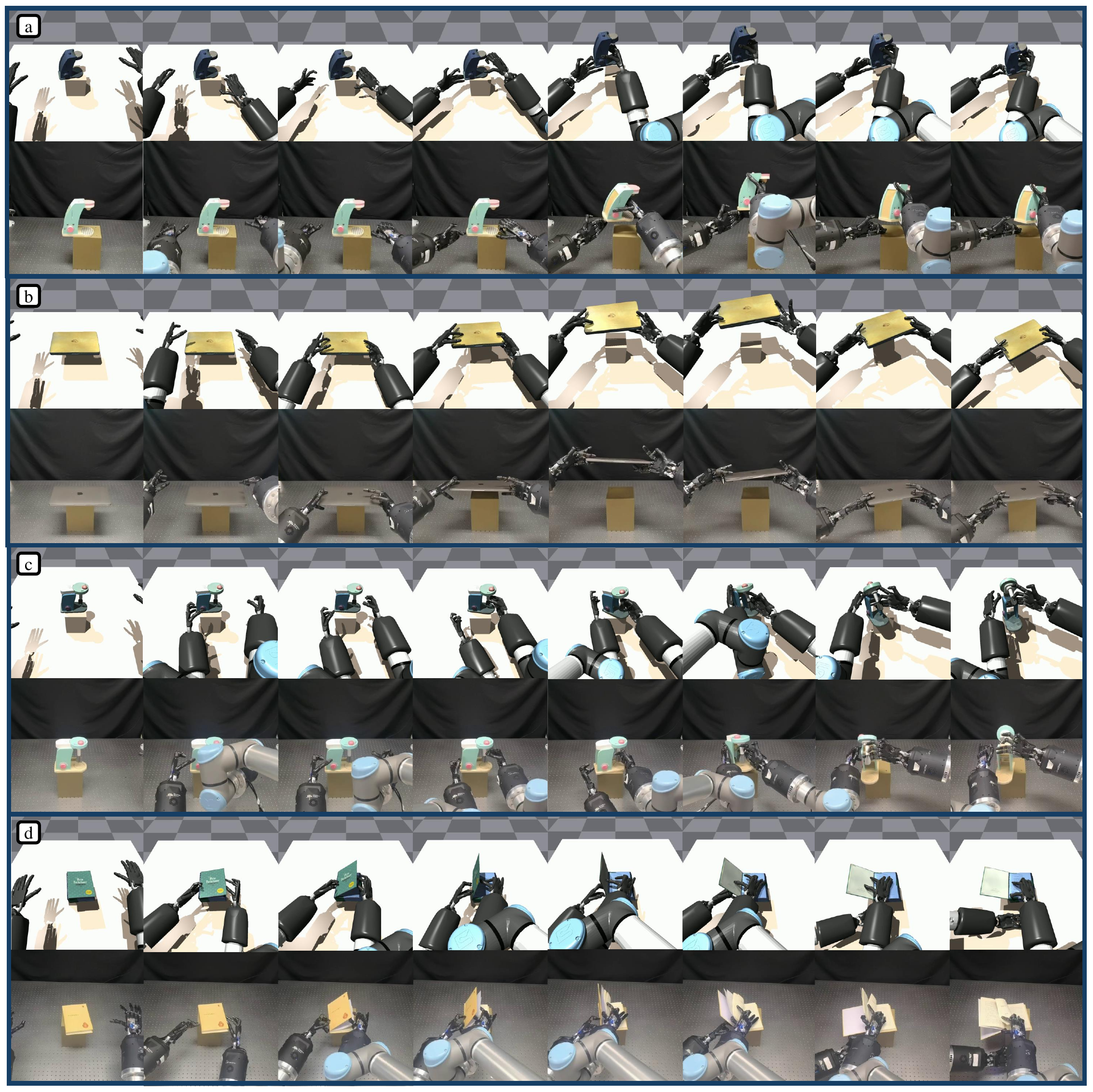}
    \caption{Real-world experiment tasks. All clips include snapshot of the simulation (top row) and the real-world (bottom row). (a) Coffee Maker : Pick and lift the coffee machine. (b) Laptop: Lift up a laptop and place it back to the table. (c) Mixer: Rotate the Mixer and then open it. (d) Notebook: Open the notebook on the table. Please refer to our website for more visualization results.}
    \label{fig:snapshot}
    \vspace{-0.4cm}
\end{figure*}

\begin{table}[t]
\centering
\small
\begin{tabular}{l|cccccc}
\toprule
 % &        & Finger & Joint & Vanilla & Ours        & Ours & Ours \\
 %  &        & IK & IK & RL & (w. FR)       & (w.o. PCIL)    &

& \multirow{2}{*}{\shortstack{Fingertip \\ Mapping}} & \multirow{2}{*}{\shortstack{Finger Joint \\ Mapping}} & \multirow{2}{*}{\shortstack{Vanilla \\RL}} & \multirow{2}{*}{\shortstack{Ours \\ (w. FR)}} & \multirow{2}{*}{\shortstack{Ours \\ (w.o. DAL)}} & \multirow{2}{*}{Ours} \\
 &   
  \\ \midrule
% Single Obj - Trained Traj & 12.1 & 8.6 & 18.4 & 31.5 &  61.1 & $\bm{85.7}$ \\
% Single Obj - Unseen Traj & -    & -    & 20.3    & -               & 44.1 & $\bm{60.6}$ \\ \midrule 
% Multi Obj - Trained Obj& 8.3       & 6.3  & 19.4 & 17.2 &  37.1       & $\bm{51.2}$ \\
%  Multi Obj - Unseen Obj& -   & -    & 10.2    & -       & 20.2       & $\bm{38.6}$ \\
Single Obj - Trained Traj& \pms{10.4}{2.8} & \pms{7.1}{4.8} & \pms{17.1}{9.6} & \pms{30.8}{12.1} &  \pms{59.6}{14.4} & \pms{\textbf{83.8}}{9.1} \\
Single Obj - Unseen Traj& \pms{4.8}{0.3}    & \pms{5.5}{0.8}    & \pms{18.8}{5.7}    & \pms{19.6}{10.1}               & \pms{42.5}{7.5} & \pms{\textbf{57.1}}{10.2} \\ \midrule 
Multi Obj - Trained Obj& \pms{6.9}{1.7}       & \pms{5.3}{1.2}  & \pms{17.7}{10.1} & \pms{15.5}{6.2} &  \pms{35.2}{2.8}       & \pms{\textbf{47.6}}{4.2} \\
 Multi Obj - Unseen Obj& \pms{3.2}{0.6}   & \pms{2.9}{0.2}    & \pms{8.1}{4.2}    & \pms{8.2}{5.3}       & \pms{18.6}{3.7}       & \pms{\textbf{36.4}}{5.0} \\
\bottomrule
\end{tabular}
\vspace{5pt}
\caption{Results for the generalization experiments.}
% \caption{Results for the genaralization experiments. Multi Traj means using one policy to solve the task-level retargeting of multiple trajectories, and generalizing to unseen trajectories. Multi Obj means using one policy to solve the task-level retargeting of multiple objects, and generalizing to unseen objects. The number in the table is the completion rate.}
\label{tab:table_4}
\vspace{-1.0cm}
\end{table}

\subsection{Transfer from simulation to real-world}
\label{sec:sim2real_results}

\textbf{Tasks.} We train one policy per object conditioned on a single goal trajectory in simulation and test its rollout performance on a real-world robot system. We use the completion rate as the evaluation metric.

\textbf{Results.} In Table~\ref{tab:table_2} real-world experiments, our approach has more than a 50\% completion rate improvements compared to prior methods, which barely achieve any success ($<20\%$ completion rate) on several objects. This result showcases the ability of our approach on tackling real-world bimaual dexterous manipulation tasks. The visualization of our real-world experiments is shown in Figure~\ref{fig:snapshot}.

%===============================================================================
\vspace{-0.2cm}

\section{Limitations}
\vspace{-0.2cm}

There are several limitations of our work. Firstly, our model encounters difficulties in manipulating small-size objects (e.g. scissors, ketchup, etc.). One potential direction is to make better use of contact information in human data. For example, ~\cite{wang2023physhoi} uses a contact graph in reward function to assist the reinforcement learning, which has the possibility of being incorporated into our framework. Secondly, we did not leverage the tactile information in the ARCTIC dataset during policy learning. In future work, we plan to equip our robot hands with tactile sensors and use contact information to train the robot to handle more contact-rich manipulation tasks. Thirdly, our method sometimes fails when the difference between the initial pose of the real-world and the goal trajectory is large. Another failure is due to the dexterous hand forms a large occlusion of the object, resulting in inaccurate pose estimation in the real-world. Fourth, The order of joints must be predefined. This limitation may not pose significant issues with single revolute joints. However, complications arise when objects feature multiple revolute joints, as the representation lacks permutation invariance across joint orders. We addressed this by focusing on category-level joint inputs consistent with URDF definitions, leveraging datasets like ShapeNet that maintain joint order consistency within categories. Future steps will involve exploring robust methods to accommodate diverse joint configurations beyond current dataset limitations. Fifth, in our high-level planner, generalizations can only happen in the same category. But we still can force the model to predict action for unseen categories by sending in a trained category id and test with unseen categories. This is how we did for the generalization experiments by using the trained Box policy to manipulate the Laptop. Since there are only 11 objects in the ARCTIC dataset, it is difficult to test the capability of the generalizations across categories. In future works we will use larger 3D object dataset and human motion capture dataset to test the generalization to unseen categories.

\vspace{-0.2cm}

\section{Conclusion}
\label{sec:conclusion}
\vspace{-0.2cm}

In this work, we present a hierarchical policy learning framework that effectively utilizes human hand motion data to train object-centric dexterous robot manipulation. At the core of our method is a high-level trajectory generative model trained with a large-scale human hand motion capture dataset, which synthesizes human-like wrist motions conditioned on the object goal trajectory. Guided by these wrist motions, we further trained an RL-based low-level finger controller to achieve the task goal. Our approach demonstrated superior performance across various household objects and showcased generalization capabilities to novel object geometries and goal trajectories. Moreover, the successful transfer of the learned policies from simulation to a real-world bimanual dexterous robot system underscores the practical applicability of our method in real-world scenarios.

\clearpage
% The acknowledgments are automatically included only in the final and preprint versions of the paper.
% \acknowledgments{If a paper is accepted, the final camera-ready version will (and probably should) include acknowledgments. All acknowledgments go at the end of the paper, including thanks to reviewers who gave useful comments, to colleagues who contributed to the ideas, and to funding agencies and corporate sponsors that provided financial support.}

%===============================================================================

\bibliography{reference}  % .bib

\begin{thebibliography}{77}
\providecommand{\natexlab}[1]{#1}
\providecommand{\url}[1]{\texttt{#1}}
\expandafter\ifx\csname urlstyle\endcsname\relax
  \providecommand{\doi}[1]{doi: #1}\else
  \providecommand{\doi}{doi: \begingroup \urlstyle{rm}\Url}\fi

\bibitem[Tassa et~al.(2018)Tassa, Doron, Muldal, Erez, Li, Casas, Budden, Abdolmaleki, Merel, Lefrancq, et~al.]{tassa2018deepmind}
Y.~Tassa, Y.~Doron, A.~Muldal, T.~Erez, Y.~Li, D.~d.~L. Casas, D.~Budden, A.~Abdolmaleki, J.~Merel, A.~Lefrancq, et~al.
\newblock Deepmind control suite.
\newblock \emph{arXiv preprint arXiv:1801.00690}, 2018.

\bibitem[James et~al.(2020)James, Ma, Arrojo, and Davison]{james2020rlbench}
S.~James, Z.~Ma, D.~R. Arrojo, and A.~J. Davison.
\newblock Rlbench: The robot learning benchmark \& learning environment.
\newblock \emph{IEEE Robotics and Automation Letters}, 5\penalty0 (2):\penalty0 3019--3026, 2020.

\bibitem[Garrett et~al.(2018)Garrett, Lozano-Perez, and Kaelbling]{garrett2018ffrob}
C.~R. Garrett, T.~Lozano-Perez, and L.~P. Kaelbling.
\newblock Ffrob: Leveraging symbolic planning for efficient task and motion planning.
\newblock \emph{The International Journal of Robotics Research}, 37\penalty0 (1):\penalty0 104--136, 2018.

\bibitem[Garrett et~al.(2021)Garrett, Chitnis, Holladay, Kim, Silver, Kaelbling, and Lozano-P{\'e}rez]{garrett2021integrated}
C.~R. Garrett, R.~Chitnis, R.~Holladay, B.~Kim, T.~Silver, L.~P. Kaelbling, and T.~Lozano-P{\'e}rez.
\newblock Integrated task and motion planning.
\newblock \emph{Annual review of control, robotics, and autonomous systems}, 4:\penalty0 265--293, 2021.

\bibitem[Huang et~al.(2022)Huang, Abbeel, Pathak, and Mordatch]{huang2022language}
W.~Huang, P.~Abbeel, D.~Pathak, and I.~Mordatch.
\newblock Language models as zero-shot planners: Extracting actionable knowledge for embodied agents.
\newblock In \emph{International Conference on Machine Learning}, pages 9118--9147. PMLR, 2022.

\bibitem[Brohan et~al.(2023)Brohan, Chebotar, Finn, Hausman, Herzog, Ho, Ibarz, Irpan, Jang, Julian, et~al.]{brohan2023can}
A.~Brohan, Y.~Chebotar, C.~Finn, K.~Hausman, A.~Herzog, D.~Ho, J.~Ibarz, A.~Irpan, E.~Jang, R.~Julian, et~al.
\newblock Do as i can, not as i say: Grounding language in robotic affordances.
\newblock In \emph{Conference on robot learning}, pages 287--318. PMLR, 2023.

\bibitem[Huang et~al.(2023)Huang, Chen, Wang, Qin, Yang, Atanasov, and Wang]{huang2023dynamic}
B.~Huang, Y.~Chen, T.~Wang, Y.~Qin, Y.~Yang, N.~Atanasov, and X.~Wang.
\newblock Dynamic handover: Throw and catch with bimanual hands.
\newblock \emph{arXiv preprint arXiv:2309.05655}, 2023.

\bibitem[Chen et~al.(2022)Chen, Wu, Wang, Feng, Jiang, Lu, McAleer, Dong, Zhu, and Yang]{chen2022towards}
Y.~Chen, T.~Wu, S.~Wang, X.~Feng, J.~Jiang, Z.~Lu, S.~McAleer, H.~Dong, S.-C. Zhu, and Y.~Yang.
\newblock Towards human-level bimanual dexterous manipulation with reinforcement learning.
\newblock \emph{Advances in Neural Information Processing Systems}, 35:\penalty0 5150--5163, 2022.

\bibitem[Zhang et~al.(2023)Zhang, Clegg, Ha, Turk, and Ye]{zhang2023learning}
Y.~Zhang, A.~Clegg, S.~Ha, G.~Turk, and Y.~Ye.
\newblock Learning to transfer in-hand manipulations using a greedy shape curriculum.
\newblock In \emph{Computer Graphics Forum}, volume~42, pages 25--36. Wiley Online Library, 2023.

\bibitem[Yin et~al.(2023)Yin, Huang, Qin, Chen, and Wang]{yin2023rotating}
Z.-H. Yin, B.~Huang, Y.~Qin, Q.~Chen, and X.~Wang.
\newblock Rotating without seeing: Towards in-hand dexterity through touch.
\newblock \emph{arXiv preprint arXiv:2303.10880}, 2023.

\bibitem[Chen et~al.(2022)Chen, Tippur, Wu, Kumar, Adelson, and Agrawal]{chen2022visual}
T.~Chen, M.~Tippur, S.~Wu, V.~Kumar, E.~Adelson, and P.~Agrawal.
\newblock Visual dexterity: In-hand dexterous manipulation from depth.
\newblock \emph{arXiv preprint arXiv:2211.11744}, 2022.

\bibitem[OpenAI et~al.(2019)OpenAI, Akkaya, Andrychowicz, Chociej, Litwin, McGrew, Petron, Paino, Plappert, Powell, Ribas, Schneider, Tezak, Tworek, Welinder, Weng, Yuan, Zaremba, and Zhang]{openai2019solving}
OpenAI, I.~Akkaya, M.~Andrychowicz, M.~Chociej, M.~Litwin, B.~McGrew, A.~Petron, A.~Paino, M.~Plappert, G.~Powell, R.~Ribas, J.~Schneider, N.~Tezak, J.~Tworek, P.~Welinder, L.~Weng, Q.~Yuan, W.~Zaremba, and L.~Zhang.
\newblock Solving rubik's cube with a robot hand.
\newblock \emph{CoRR}, abs/1910.07113, 2019.

\bibitem[Fan et~al.(2023)Fan, Taheri, Tzionas, Kocabas, Kaufmann, Black, and Hilliges]{fan2023arctic}
Z.~Fan, O.~Taheri, D.~Tzionas, M.~Kocabas, M.~Kaufmann, M.~J. Black, and O.~Hilliges.
\newblock Arctic: A dataset for dexterous bimanual hand-object manipulation.
\newblock In \emph{Proceedings of the IEEE/CVF Conference on Computer Vision and Pattern Recognition}, pages 12943--12954, 2023.

\bibitem[Bai and Liu(2014)]{bai2014dexterous}
Y.~Bai and C.~K. Liu.
\newblock Dexterous manipulation using both palm and fingers.
\newblock In \emph{2014 IEEE International Conference on Robotics and Automation (ICRA)}, pages 1560--1565. IEEE, 2014.

\bibitem[Kumar et~al.(2014)Kumar, Tassa, Erez, and Todorov]{kumar2014real}
V.~Kumar, Y.~Tassa, T.~Erez, and E.~Todorov.
\newblock Real-time behaviour synthesis for dynamic hand-manipulation.
\newblock In \emph{2014 IEEE International Conference on Robotics and Automation (ICRA)}, pages 6808--6815. IEEE, 2014.

\bibitem[Mason and Salisbury~Jr(1985)]{mason1985robot}
M.~T. Mason and J.~K. Salisbury~Jr.
\newblock Robot hands and the mechanics of manipulation.
\newblock 1985.

\bibitem[Mordatch et~al.(2012)Mordatch, Popovi{\'c}, and Todorov]{mordatch2012contact}
I.~Mordatch, Z.~Popovi{\'c}, and E.~Todorov.
\newblock Contact-invariant optimization for hand manipulation.
\newblock In \emph{Proceedings of the ACM SIGGRAPH/Eurographics symposium on computer animation}, pages 137--144, 2012.

\bibitem[Chen et~al.(2024)Chen, Bohg, and Liu]{chen2024springgrasp}
S.~Chen, J.~Bohg, and C.~K. Liu.
\newblock Springgrasp: An optimization pipeline for robust and compliant dexterous pre-grasp synthesis.
\newblock \emph{arXiv preprint arXiv:2404.13532}, 2024.

\bibitem[Pang et~al.(2023)Pang, Suh, Yang, and Tedrake]{pang2023global}
T.~Pang, H.~T. Suh, L.~Yang, and R.~Tedrake.
\newblock Global planning for contact-rich manipulation via local smoothing of quasi-dynamic contact models.
\newblock \emph{IEEE Transactions on robotics}, 2023.

\bibitem[Cheng et~al.(2023)Cheng, Patil, Temel, Kroemer, and Mason]{cheng2023enhancing}
X.~Cheng, S.~Patil, Z.~Temel, O.~Kroemer, and M.~T. Mason.
\newblock Enhancing dexterity in robotic manipulation via hierarchical contact exploration.
\newblock \emph{IEEE Robotics and Automation Letters}, 9\penalty0 (1):\penalty0 390--397, 2023.

\bibitem[Chen et~al.(2021)Chen, Xu, and Agrawal]{chen2021system}
T.~Chen, J.~Xu, and P.~Agrawal.
\newblock A system for general in-hand object re-orientation.
\newblock \emph{Conference on Robot Learning}, 2021.

\bibitem[Qi et~al.(2023{\natexlab{a}})Qi, Yi, Suresh, Lambeta, Ma, Calandra, and Malik]{qi2023general}
H.~Qi, B.~Yi, S.~Suresh, M.~Lambeta, Y.~Ma, R.~Calandra, and J.~Malik.
\newblock General in-hand object rotation with vision and touch.
\newblock In \emph{Conference on Robot Learning}, pages 2549--2564. PMLR, 2023{\natexlab{a}}.

\bibitem[Qi et~al.(2023{\natexlab{b}})Qi, Kumar, Calandra, Ma, and Malik]{qi2023hand}
H.~Qi, A.~Kumar, R.~Calandra, Y.~Ma, and J.~Malik.
\newblock In-hand object rotation via rapid motor adaptation.
\newblock In \emph{Conference on Robot Learning}, pages 1722--1732. PMLR, 2023{\natexlab{b}}.

\bibitem[Dasari et~al.(2023)Dasari, Gupta, and Kumar]{dasari2023learning}
S.~Dasari, A.~Gupta, and V.~Kumar.
\newblock Learning dexterous manipulation from exemplar object trajectories and pre-grasps.
\newblock In \emph{2023 IEEE International Conference on Robotics and Automation (ICRA)}, pages 3889--3896. IEEE, 2023.

\bibitem[Yang et~al.(2024)Yang, Lu, Church, Lin, Ford, Li, Psomopoulou, Barton, and Lepora]{yang2024anyrotate}
M.~Yang, C.~Lu, A.~Church, Y.~Lin, C.~Ford, H.~Li, E.~Psomopoulou, D.~A. Barton, and N.~F. Lepora.
\newblock Anyrotate: Gravity-invariant in-hand object rotation with sim-to-real touch.
\newblock \emph{arXiv preprint arXiv:2405.07391}, 2024.

\bibitem[Pitz et~al.(2023)Pitz, R{\"o}stel, Sievers, and B{\"a}uml]{pitz2023dextrous}
J.~Pitz, L.~R{\"o}stel, L.~Sievers, and B.~B{\"a}uml.
\newblock Dextrous tactile in-hand manipulation using a modular reinforcement learning architecture.
\newblock In \emph{2023 IEEE International Conference on Robotics and Automation (ICRA)}, pages 1852--1858. IEEE, 2023.

\bibitem[Handa et~al.(2022)Handa, Allshire, Makoviychuk, Petrenko, Singh, Liu, Makoviichuk, Van~Wyk, Zhurkevich, Sundaralingam, et~al.]{handa2022dextreme}
A.~Handa, A.~Allshire, V.~Makoviychuk, A.~Petrenko, R.~Singh, J.~Liu, D.~Makoviichuk, K.~Van~Wyk, A.~Zhurkevich, B.~Sundaralingam, et~al.
\newblock Dextreme: Transfer of agile in-hand manipulation from simulation to reality.
\newblock \emph{arXiv preprint arXiv:2210.13702}, 2022.

\bibitem[Khandate et~al.(2023)Khandate, Shang, Chang, Saidi, Liu, Dennis, Adams, and Ciocarlie]{khandate2023sampling}
G.~Khandate, S.~Shang, E.~T. Chang, T.~L. Saidi, Y.~Liu, S.~M. Dennis, J.~Adams, and M.~Ciocarlie.
\newblock Sampling-based exploration for reinforcement learning of dexterous manipulation.
\newblock \emph{arXiv preprint arXiv:2303.03486}, 2023.

\bibitem[Lin et~al.(2024)Lin, Yin, Qi, Abbeel, and Malik]{lin2024twisting}
T.~Lin, Z.-H. Yin, H.~Qi, P.~Abbeel, and J.~Malik.
\newblock Twisting lids off with two hands.
\newblock \emph{arXiv preprint arXiv:2403.02338}, 2024.

\bibitem[Xu et~al.(2023)Xu, Hu, Doshi, Rovinsky, Kumar, Gupta, and Levine]{xu2023dexterous}
K.~Xu, Z.~Hu, R.~Doshi, A.~Rovinsky, V.~Kumar, A.~Gupta, and S.~Levine.
\newblock Dexterous manipulation from images: Autonomous real-world rl via substep guidance.
\newblock In \emph{2023 IEEE International Conference on Robotics and Automation (ICRA)}, pages 5938--5945. IEEE, 2023.

\bibitem[Chen et~al.(2023)Chen, Wang, Fei-Fei, and Liu]{chen2023sequential}
Y.~Chen, C.~Wang, L.~Fei-Fei, and C.~K. Liu.
\newblock Sequential dexterity: Chaining dexterous policies for long-horizon manipulation.
\newblock \emph{arXiv preprint arXiv:2309.00987}, 2023.

\bibitem[Gupta et~al.(2021)Gupta, Yu, Zhao, Kumar, Rovinsky, Xu, Devlin, and Levine]{gupta2021reset}
A.~Gupta, J.~Yu, T.~Z. Zhao, V.~Kumar, A.~Rovinsky, K.~Xu, T.~Devlin, and S.~Levine.
\newblock Reset-free reinforcement learning via multi-task learning: Learning dexterous manipulation behaviors without human intervention.
\newblock In \emph{2021 IEEE International Conference on Robotics and Automation (ICRA)}, pages 6664--6671. IEEE, 2021.

\bibitem[Zakka et~al.(2023)Zakka, Smith, Gileadi, Howell, Peng, Singh, Tassa, Florence, Zeng, and Abbeel]{zakka2023robopianist}
K.~Zakka, L.~Smith, N.~Gileadi, T.~Howell, X.~B. Peng, S.~Singh, Y.~Tassa, P.~Florence, A.~Zeng, and P.~Abbeel.
\newblock Robopianist: A benchmark for high-dimensional robot control.
\newblock \emph{arXiv preprint arXiv:2304.04150}, 2023.

\bibitem[Rajeswaran et~al.(2017)Rajeswaran, Kumar, Gupta, Vezzani, Schulman, Todorov, and Levine]{rajeswaran2017learning}
A.~Rajeswaran, V.~Kumar, A.~Gupta, G.~Vezzani, J.~Schulman, E.~Todorov, and S.~Levine.
\newblock Learning complex dexterous manipulation with deep reinforcement learning and demonstrations.
\newblock \emph{arXiv preprint arXiv:1709.10087}, 2017.

\bibitem[Radosavovic et~al.(2021)Radosavovic, Wang, Pinto, and Malik]{radosavovic2021state}
I.~Radosavovic, X.~Wang, L.~Pinto, and J.~Malik.
\newblock State-only imitation learning for dexterous manipulation.
\newblock In \emph{2021 IEEE/RSJ International Conference on Intelligent Robots and Systems (IROS)}, pages 7865--7871. IEEE, 2021.

\bibitem[Arunachalam et~al.(2023)Arunachalam, G{\"u}zey, Chintala, and Pinto]{arunachalam2023holo}
S.~P. Arunachalam, I.~G{\"u}zey, S.~Chintala, and L.~Pinto.
\newblock Holo-dex: Teaching dexterity with immersive mixed reality.
\newblock In \emph{2023 IEEE International Conference on Robotics and Automation (ICRA)}, pages 5962--5969. IEEE, 2023.

\bibitem[Guzey et~al.(2023)Guzey, Evans, Chintala, and Pinto]{guzey2023dexterity}
I.~Guzey, B.~Evans, S.~Chintala, and L.~Pinto.
\newblock Dexterity from touch: Self-supervised pre-training of tactile representations with robotic play.
\newblock \emph{arXiv preprint arXiv:2303.12076}, 2023.

\bibitem[Handa et~al.(2020)Handa, Van~Wyk, Yang, Liang, Chao, Wan, Birchfield, Ratliff, and Fox]{handa2020dexpilot}
A.~Handa, K.~Van~Wyk, W.~Yang, J.~Liang, Y.-W. Chao, Q.~Wan, S.~Birchfield, N.~Ratliff, and D.~Fox.
\newblock Dexpilot: Vision-based teleoperation of dexterous robotic hand-arm system.
\newblock In \emph{2020 IEEE International Conference on Robotics and Automation (ICRA)}, pages 9164--9170. IEEE, 2020.

\bibitem[Sivakumar et~al.(2022)Sivakumar, Shaw, and Pathak]{sivakumar2022robotic}
A.~Sivakumar, K.~Shaw, and D.~Pathak.
\newblock Robotic telekinesis: Learning a robotic hand imitator by watching humans on youtube.
\newblock \emph{arXiv preprint arXiv:2202.10448}, 2022.

\bibitem[Qin et~al.(2023)Qin, Yang, Huang, Van~Wyk, Su, Wang, Chao, and Fox]{qin2023anyteleop}
Y.~Qin, W.~Yang, B.~Huang, K.~Van~Wyk, H.~Su, X.~Wang, Y.-W. Chao, and D.~Fox.
\newblock Anyteleop: A general vision-based dexterous robot arm-hand teleoperation system.
\newblock \emph{arXiv preprint arXiv:2307.04577}, 2023.

\bibitem[Mandikal and Grauman(2021)]{mandikal2021learning}
P.~Mandikal and K.~Grauman.
\newblock Learning dexterous grasping with object-centric visual affordances.
\newblock In \emph{2021 IEEE international conference on robotics and automation (ICRA)}, pages 6169--6176. IEEE, 2021.

\bibitem[Chen et~al.(2022)Chen, Van~Wyk, Chao, Yang, Mousavian, Gupta, and Fox]{chen2022learning}
Z.~Q. Chen, K.~Van~Wyk, Y.-W. Chao, W.~Yang, A.~Mousavian, A.~Gupta, and D.~Fox.
\newblock Learning robust real-world dexterous grasping policies via implicit shape augmentation.
\newblock \emph{arXiv preprint arXiv:2210.13638}, 2022.

\bibitem[Qin et~al.(2022)Qin, Wu, Liu, Jiang, Yang, Fu, and Wang]{qin2022dexmv}
Y.~Qin, Y.-H. Wu, S.~Liu, H.~Jiang, R.~Yang, Y.~Fu, and X.~Wang.
\newblock Dexmv: Imitation learning for dexterous manipulation from human videos.
\newblock In \emph{European Conference on Computer Vision}, pages 570--587. Springer, 2022.

\bibitem[Cui et~al.(2022)Cui, Wang, Shafiullah, and Pinto]{cui2022play}
Z.~J. Cui, Y.~Wang, N.~M.~M. Shafiullah, and L.~Pinto.
\newblock From play to policy: Conditional behavior generation from uncurated robot data.
\newblock \emph{arXiv e-prints}, pages arXiv--2210, 2022.

\bibitem[Haldar et~al.(2023)Haldar, Pari, Rai, and Pinto]{haldar2023teach}
S.~Haldar, J.~Pari, A.~Rai, and L.~Pinto.
\newblock Teach a robot to fish: Versatile imitation from one minute of demonstrations.
\newblock \emph{arXiv preprint arXiv:2303.01497}, 2023.

\bibitem[Qin et~al.(2022)Qin, Su, and Wang]{qin2022one}
Y.~Qin, H.~Su, and X.~Wang.
\newblock From one hand to multiple hands: Imitation learning for dexterous manipulation from single-camera teleoperation.
\newblock \emph{IEEE Robotics and Automation Letters}, 7\penalty0 (4):\penalty0 10873--10881, 2022.

\bibitem[Arunachalam et~al.(2022)Arunachalam, Silwal, Evans, and Pinto]{arunachalam2022dexterous}
S.~P. Arunachalam, S.~Silwal, B.~Evans, and L.~Pinto.
\newblock Dexterous imitation made easy: A learning-based framework for efficient dexterous manipulation.
\newblock \emph{arXiv preprint arXiv:2203.13251}, 2022.

\bibitem[Guzey et~al.(2023)Guzey, Dai, Evans, Chintala, and Pinto]{guzey2023see}
I.~Guzey, Y.~Dai, B.~Evans, S.~Chintala, and L.~Pinto.
\newblock See to touch: Learning tactile dexterity through visual incentives.
\newblock \emph{arXiv preprint arXiv:2309.12300}, 2023.

\bibitem[Lin et~al.(2024)Lin, Zhang, Li, Qi, Yi, Levine, and Malik]{lin2024learning}
T.~Lin, Y.~Zhang, Q.~Li, H.~Qi, B.~Yi, S.~Levine, and J.~Malik.
\newblock Learning visuotactile skills with two multifingered hands.
\newblock \emph{arXiv preprint arXiv:2404.16823}, 2024.

\bibitem[Ze et~al.(2024)Ze, Zhang, Zhang, Hu, Wang, and Xu]{ze20243d}
Y.~Ze, G.~Zhang, K.~Zhang, C.~Hu, M.~Wang, and H.~Xu.
\newblock 3d diffusion policy.
\newblock \emph{arXiv preprint arXiv:2403.03954}, 2024.

\bibitem[Bahl et~al.(2022)Bahl, Gupta, and Pathak]{bahl2022human}
S.~Bahl, A.~Gupta, and D.~Pathak.
\newblock Human-to-robot imitation in the wild.
\newblock \emph{arXiv preprint arXiv:2207.09450}, 2022.

\bibitem[Mandikal and Grauman(2022)]{mandikal2022dexvip}
P.~Mandikal and K.~Grauman.
\newblock Dexvip: Learning dexterous grasping with human hand pose priors from video.
\newblock In \emph{Conference on Robot Learning}, pages 651--661. PMLR, 2022.

\bibitem[Bahl et~al.(2023)Bahl, Mendonca, Chen, Jain, and Pathak]{bahl2023affordances}
S.~Bahl, R.~Mendonca, L.~Chen, U.~Jain, and D.~Pathak.
\newblock Affordances from human videos as a versatile representation for robotics.
\newblock In \emph{Proceedings of the IEEE/CVF Conference on Computer Vision and Pattern Recognition}, pages 13778--13790, 2023.

\bibitem[Bharadhwaj et~al.(2023)Bharadhwaj, Gupta, Kumar, and Tulsiani]{bharadhwaj2023towards}
H.~Bharadhwaj, A.~Gupta, V.~Kumar, and S.~Tulsiani.
\newblock Towards generalizable zero-shot manipulation via translating human interaction plans.
\newblock \emph{arXiv preprint arXiv:2312.00775}, 2023.

\bibitem[Shaw et~al.(2023)Shaw, Bahl, and Pathak]{shaw2023videodex}
K.~Shaw, S.~Bahl, and D.~Pathak.
\newblock Videodex: Learning dexterity from internet videos.
\newblock In \emph{Conference on Robot Learning}, pages 654--665. PMLR, 2023.

\bibitem[Bahety et~al.(2024)Bahety, Mandikal, Abbatematteo, and Mart{\'\i}n-Mart{\'\i}n]{bahety2024screwmimic}
A.~Bahety, P.~Mandikal, B.~Abbatematteo, and R.~Mart{\'\i}n-Mart{\'\i}n.
\newblock Screwmimic: Bimanual imitation from human videos with screw space projection.
\newblock \emph{arXiv preprint arXiv:2405.03666}, 2024.

\bibitem[Zhang et~al.(2023)Zhang, Christen, Fan, Zheng, Hwangbo, Song, and Hilliges]{zhang2023artigrasp}
H.~Zhang, S.~Christen, Z.~Fan, L.~Zheng, J.~Hwangbo, J.~Song, and O.~Hilliges.
\newblock Artigrasp: Physically plausible synthesis of bi-manual dexterous grasping and articulation.
\newblock \emph{arXiv preprint arXiv:2309.03891}, 2023.

\bibitem[Chen et~al.(2022)Chen, Van~Wyk, Chao, Yang, Mousavian, Gupta, and Fox]{chen2022dextransfer}
Z.~Q. Chen, K.~Van~Wyk, Y.-W. Chao, W.~Yang, A.~Mousavian, A.~Gupta, and D.~Fox.
\newblock Dextransfer: Real world multi-fingered dexterous grasping with minimal human demonstrations.
\newblock \emph{arXiv preprint arXiv:2209.14284}, 2022.

\bibitem[Wang et~al.(2023)Wang, Fan, Sun, Zhang, Fei-Fei, Xu, Zhu, and Anandkumar]{wang2023mimicplay}
C.~Wang, L.~Fan, J.~Sun, R.~Zhang, L.~Fei-Fei, D.~Xu, Y.~Zhu, and A.~Anandkumar.
\newblock Mimicplay: Long-horizon imitation learning by watching human play.
\newblock \emph{arXiv preprint arXiv:2302.12422}, 2023.

\bibitem[Garcia-Hernando et~al.(2018)Garcia-Hernando, Yuan, Baek, and Kim]{garcia2018first}
G.~Garcia-Hernando, S.~Yuan, S.~Baek, and T.-K. Kim.
\newblock First-person hand action benchmark with rgb-d videos and 3d hand pose annotations.
\newblock In \emph{Proceedings of the IEEE conference on computer vision and pattern recognition}, pages 409--419, 2018.

\bibitem[Hasson et~al.(2019)Hasson, Varol, Tzionas, Kalevatykh, Black, Laptev, and Schmid]{hasson2019learning}
Y.~Hasson, G.~Varol, D.~Tzionas, I.~Kalevatykh, M.~J. Black, I.~Laptev, and C.~Schmid.
\newblock Learning joint reconstruction of hands and manipulated objects.
\newblock In \emph{Proceedings of the IEEE/CVF conference on computer vision and pattern recognition}, pages 11807--11816, 2019.

\bibitem[Taheri et~al.(2020)Taheri, Ghorbani, Black, and Tzionas]{taheri2020grab}
O.~Taheri, N.~Ghorbani, M.~J. Black, and D.~Tzionas.
\newblock Grab: A dataset of whole-body human grasping of objects.
\newblock In \emph{Computer Vision--ECCV 2020: 16th European Conference, Glasgow, UK, August 23--28, 2020, Proceedings, Part IV 16}, pages 581--600. Springer, 2020.

\bibitem[Wang et~al.(2024)Wang, Shi, Wang, Zhang, Fei-Fei, and Liu]{wang2024dexcap}
C.~Wang, H.~Shi, W.~Wang, R.~Zhang, L.~Fei-Fei, and C.~K. Liu.
\newblock Dexcap: Scalable and portable mocap data collection system for dexterous manipulation.
\newblock \emph{arXiv preprint arXiv:2403.07788}, 2024.

\bibitem[Ye and Liu(2012)]{ye2012synthesis}
Y.~Ye and C.~K. Liu.
\newblock Synthesis of detailed hand manipulations using contact sampling.
\newblock \emph{ACM Transactions on Graphics (ToG)}, 31\penalty0 (4):\penalty0 1--10, 2012.

\bibitem[Ho and Ermon(2016)]{ho2016generative}
J.~Ho and S.~Ermon.
\newblock Generative adversarial imitation learning.
\newblock \emph{Advances in neural information processing systems}, 29, 2016.

\bibitem[Peng et~al.(2018)Peng, Abbeel, Levine, and Van~de Panne]{peng2018deepmimic}
X.~B. Peng, P.~Abbeel, S.~Levine, and M.~Van~de Panne.
\newblock Deepmimic: Example-guided deep reinforcement learning of physics-based character skills.
\newblock \emph{ACM Transactions On Graphics (TOG)}, 37\penalty0 (4):\penalty0 1--14, 2018.

\bibitem[Peng et~al.(2021)Peng, Ma, Abbeel, Levine, and Kanazawa]{peng2021amp}
X.~B. Peng, Z.~Ma, P.~Abbeel, S.~Levine, and A.~Kanazawa.
\newblock Amp: Adversarial motion priors for stylized physics-based character control.
\newblock \emph{ACM Transactions on Graphics (ToG)}, 40\penalty0 (4):\penalty0 1--20, 2021.

\bibitem[Peng et~al.(2022)Peng, Guo, Halper, Levine, and Fidler]{peng2022ase}
X.~B. Peng, Y.~Guo, L.~Halper, S.~Levine, and S.~Fidler.
\newblock Ase: Large-scale reusable adversarial skill embeddings for physically simulated characters.
\newblock \emph{ACM Transactions On Graphics (TOG)}, 41\penalty0 (4):\penalty0 1--17, 2022.

\bibitem[Wang et~al.(2023)Wang, Lin, Zeng, Luo, Zhang, and Zhang]{wang2023physhoi}
Y.~Wang, J.~Lin, A.~Zeng, Z.~Luo, J.~Zhang, and L.~Zhang.
\newblock Physhoi: Physics-based imitation of dynamic human-object interaction.
\newblock \emph{arXiv preprint arXiv:2312.04393}, 2023.

\bibitem[Schulman et~al.(2017)Schulman, Wolski, Dhariwal, Radford, and Klimov]{schulman2017proximal}
J.~Schulman, F.~Wolski, P.~Dhariwal, A.~Radford, and O.~Klimov.
\newblock Proximal policy optimization algorithms.
\newblock \emph{arXiv preprint arXiv:1707.06347}, 2017.

\bibitem[Rusu et~al.(2016)Rusu, Colmenarejo, G{\"{u}}l{\c{c}}ehre, Desjardins, Kirkpatrick, Pascanu, Mnih, Kavukcuoglu, and Hadsell]{rusu2015policy}
A.~A. Rusu, S.~G. Colmenarejo, {\c{C}}.~G{\"{u}}l{\c{c}}ehre, G.~Desjardins, J.~Kirkpatrick, R.~Pascanu, V.~Mnih, K.~Kavukcuoglu, and R.~Hadsell.
\newblock Policy distillation.
\newblock In \emph{{ICLR} (Poster)}, 2016.

\bibitem[Wen et~al.(2023)Wen, Yang, Kautz, and Birchfield]{wen2023foundationpose}
B.~Wen, W.~Yang, J.~Kautz, and S.~Birchfield.
\newblock Foundationpose: Unified 6d pose estimation and tracking of novel objects.
\newblock \emph{arXiv preprint arXiv:2312.08344}, 2023.

\bibitem[Yu et~al.(2020)Yu, Quillen, He, Julian, Hausman, Finn, and Levine]{yu2020meta}
T.~Yu, D.~Quillen, Z.~He, R.~Julian, K.~Hausman, C.~Finn, and S.~Levine.
\newblock Meta-world: A benchmark and evaluation for multi-task and meta reinforcement learning.
\newblock In \emph{Conference on robot learning}, pages 1094--1100. PMLR, 2020.

\bibitem[Charlesworth and Montana(2021)]{charlesworth2021solving}
H.~J. Charlesworth and G.~Montana.
\newblock Solving challenging dexterous manipulation tasks with trajectory optimisation and reinforcement learning.
\newblock In \emph{International Conference on Machine Learning}, pages 1496--1506. PMLR, 2021.

\bibitem[Ross et~al.(2011)Ross, Gordon, and Bagnell]{ross2011reduction}
S.~Ross, G.~Gordon, and D.~Bagnell.
\newblock A reduction of imitation learning and structured prediction to no-regret online learning.
\newblock In \emph{Proceedings of the fourteenth international conference on artificial intelligence and statistics}, pages 627--635. JMLR Workshop and Conference Proceedings, 2011.

\bibitem[Li et~al.(2023)Li, Clegg, Mottaghi, Wu, Puig, and Liu]{li2023controllable}
J.~Li, A.~Clegg, R.~Mottaghi, J.~Wu, X.~Puig, and C.~K. Liu.
\newblock Controllable human-object interaction synthesis.
\newblock \emph{arXiv preprint arXiv:2312.03913}, 2023.

\bibitem[Pezzato et~al.(2023)Pezzato, Salmi, Spahn, Trevisan, Alonso-Mora, and Corbato]{pezzato2023samplingbased}
C.~Pezzato, C.~Salmi, M.~Spahn, E.~Trevisan, J.~Alonso-Mora, and C.~H. Corbato.
\newblock Sampling-based model predictive control leveraging parallelizable physics simulations, 2023.

\end{thebibliography}
\appendix

% \tableofappendix
\newpage

\section{Data Augmentation Loop}\label{app:dal}

\subsection{Pseudo Code}

\begin{algorithm}[!h]
    \caption{\textsc{Data Augmentation Loop}}
    \label{alg:method}
    \begin{algorithmic}[1]
        \REQUIRE Human data $D = (G, a^{W})$, training set $D_t \in D$, high-level planner $\pi^H$, low-level controller $\pi^L$, augmentation iteration $L$, data augment function $L_{aug}()$, wrist pose trajectories $G_t = (\bm{g}_t,\bm{g}_{t+1},..., \bm{g}_{t+T-1})$, goal trajectory of the object and its geometric features $\bm{a}^W_t = (\bm{a}^W_t,\bm{a}^W_{t+1},..., \bm{a}^W_{t+T-1})$.
        \STATE Initialize $\pi^H$, $\pi^L$, $D_t = \{\}$.
    \FOR{iteration $m=0,1,...,L$} 
        \WHILE{until convergence of $\pi^H$}
            \STATE Generate augmented data $L_{aug}(D)$
            \STATE Append into training set $D_t \leftarrow D_t + L_{aug}(D)$
            \STATE Train $\pi^H$ on $D_t$
        \ENDWHILE            

        \WHILE{until convergence of $\pi^L$}
            \STATE Train $\pi^L$ on $(\pi^H(G), G)$
        \ENDWHILE   
        \STATE Rollout success trajectories $D_s^t = (\bm{a}^W_t, G_t)$ with $\pi^{L}$ 
        \STATE Append into human data $D \leftarrow D_T + D_s^t$
    \ENDFOR
    \end{algorithmic}
\end{algorithm}

\subsection{Detail of the Data Augmentation Loop}

% We have three types of augmentation during the RL training of the $\pi^L$: randomizing the object's mesh scales in all three dimensions (width, length, height), randomizing the object's initial pose, and modifying the goal trajectories of the object with waypoint interpolation. 
Below are the details for each augmentation. The unit of length is centimeters and the unit of angle is degrees.
\begin{itemize}[leftmargin=*]
\item Random the object's mesh scales with a small scale:
\begin{itemize}[leftmargin=*]
 \item The scale of the width of the manipulated object ranges from 0.9 to 1.1.
 \item The scale of the length of the manipulated object ranges from 0.9 to 1.1.
 \item The scale of the height of the manipulated object ranges from 0.9 to 1.1.
\end{itemize}

\item Random the object's initial pose with a small scale:
\begin{itemize}[leftmargin=*]
 \item The x-coordinate of the manipulated object ranges from -0.02 to 0.02.
 \item The y-coordinate of the manipulated object ranges from -0.02 to 0.02.
 \item The manipulated object's z-axis Euler degree ranges from 0 to 30.
\end{itemize}

\item Modify the goal trajectories of the object with waypoint interpolation:
\begin{itemize}[leftmargin=*]
 \item The x-coordinate of the goal trajectories position is added by ranges from -0.02 to 0.02.
 \item The y-coordinate of the goal trajectories position is added by ranges from -0.02 to 0.02.
 \item The z-coordinate of the goal trajectories position is added by ranges from -0.02 to 0.02.
\end{itemize}

\end{itemize}

\section{Detail Implementation of RL in Simulation}\label{app:RL_implement}

\subsection{Observation Space}

Table.\ref{app:observation_space} gives the specific information of the observation space.

% \subsection{Action Space}

% Our action space is divided into the control of the dexterous hands and the control of the arm. For the control of the hands, we directly output the target angle value of the hand joints, and move them to the target position through a PD controller. For the control of the arm, we used the Residual Action Space~\cite{johannink2019residual} as our Action Space to control the dual robot arms. The core insight of the Residual Action Space is use a trajectory as reference for the arm movement, and the RL only needs to make some fine-tuning on top of it. This is very natural in our framework, because we have learned the reference trajectory of the hand wrists from the high-level planner.

\subsection{Reward Design}

Denote the $\bm{\hat{g}}_i^R$, $\bm{\hat{g}}_i^T$ and $\hat{g}_i^J$ is the current 3D translation, 3D rotation and joint angle of the object respectively, the desired object 3D rotation $\bm{g}_i^R$, the desired object 3D translation $\bm{g}_i^T$, and the desired object joint angle $g_i^J$. $\lambda_1$, $\lambda_2$ and $\lambda_3$ is the hyperparameters to balance the weight of each component of the reward.

The reward function is defined as: 
\begin{equation}
 \begin{aligned}
r_t=\exp^{ -(\lambda_1*\Vert \bm{g}^R_t - \bm{\hat{g}}_t^R \Vert_2+\lambda_2*\Vert \bm{g}^T_t - \bm{\hat{g}}_t^T \Vert_2+\lambda_3*\Vert g^J_t - \hat{g}_t^J \Vert_2)}
 \end{aligned}
\end{equation}
where $\lambda_1 = 20$, $\lambda_2 = 1$, and $\lambda_3 = 5$.

We use an exponential map in the reward function, which is an effective reward shaping technique used in the case to minimize the distance, introduced by~\cite{yu2020meta, charlesworth2021solving}. To improve the calculation efficiency, we use quaternion to represent the object orientation. The angular position difference is then computed through the dot product between the normalized goal quaternion and the current object’s quaternion.

\section{Detail Implementation in Real-World}\label{app:real_implement}

\begin{figure*}[!t]
    \centering
    \includegraphics[width=0.9\linewidth]{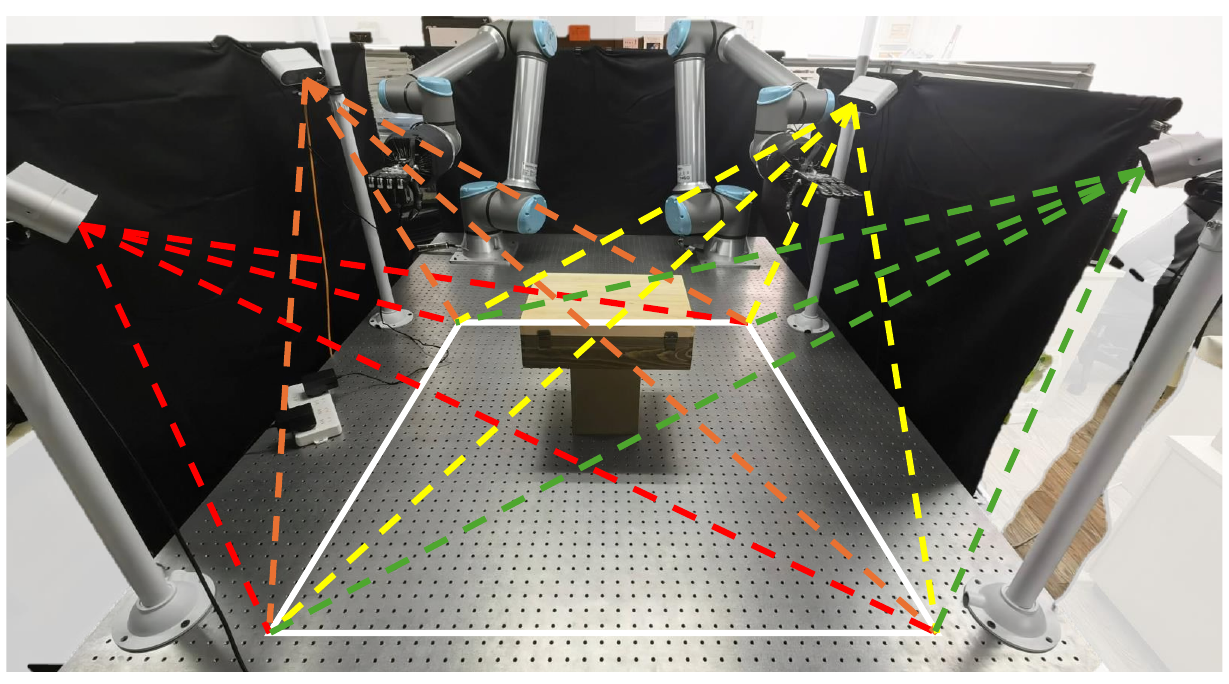}
    \caption{Setup of the cameras.}
    \label{fig:perception_setup}
\end{figure*}

\begin{table}[!t]
    \centering
    \begin{tabular}{c|c}
    \toprule
    \rowcolor[HTML]{D4F7EE}
    Index & Description\\\hline
    0 - 60 & right arm-hand dof position, velocity\\\hline
    60 - 120 & left arm-hand dof position, velocity\\\hline
    120 - 133 & right hand end-effector position, velocity, linear velocity, angle velocity\\\hline
    133 - 146 &	left hand end-effector position, velocity, linear velocity, angle velocity\\\hline
    146 - 159 &	object base position, rotation, linear velocity, angle velocity\\\hline
    159 - 172 &	articulated object top part position, rotation, linear velocity, angle velocity\\\hline
    172 - 185 &	articulated object bottom part position, rotation, linear velocity, angle velocity\\\hline
    185 - 187 &	object dof position, velocity\\\hline
    187 - 257 & desired object motion trajectory $G = (\bm{g}_t,\bm{g}_{t+1},..., \bm{g}_{t+T})$\\\hline
    257 - 397 & sequence of 6-DoF wrist actions $(\bm{a}^W_t,\bm{a}^W_{t+1},..., \bm{a}^W_{t+T})$ generated by high-level planner\\\hline
    397 - 462 & right hand fingertip pose, linear velocity, angle velocity\\\hline
    462 - 527 & left hand fingertip pose, linear velocity, angle velocity\\
    \bottomrule
    \end{tabular}
    \vspace{5pt}
    \caption{Observation space of our framework in simulation.}
    \label{app:observation_space}
\end{table}

\subsection{Perception}

Our perception setup is shown in Figure~\ref{fig:perception_setup}. We arranged 4 identical Femto Bolt cameras around the table and face towards the object. We use FoundationPose~\cite{wen2023foundationpose} to estimate the articulated object pose. To remove the abnormal results, we compare each pose to the desired pose and remove the pose if the error is smaller than a threshold ($5$ centimeters in translation and $0.5$ radians in orientation). Finally, we average the rest of the poses as our observation for the policy. If none of the poses is smaller than the threshold, we continue to use the pose from the previous frame.

\subsection{Policy Distillation}
We use the DAgger~\cite{ross2011reduction} algorithm for policy distillation. Table.\ref{app:observation_real} gives the specific information of the observation space of the distilled policy.

\begin{table}[thbp]
    \centering
    \begin{tabular}{c|c}
    \toprule
    \rowcolor[HTML]{D4F7EE}
    Index & Description\\\hline
    0 - 24 & right hand dof position\\\hline
    24 - 48 & left hand dof position\\\hline
    48 - 55 & right hand end-effector position, rotation\\\hline
    55 - 62 &	left hand end-effector position, rotation\\\hline
    62 - 69 &	articulated object top part position, rotation\\\hline
    69 - 76 &	articulated object bottom part position, rotation\\\hline
    76 - 77 &	object dof position\\\hline
    77 - 147 & desired object motion trajectory $G = (\bm{g}_t,\bm{g}_{t+1},..., \bm{g}_{t+T})$\\\hline
    147 - 287 & sequence of 6-DoF wrist actions $(\bm{a}^W_t,\bm{a}^W_{t+1},..., \bm{a}^W_{t+T})$ generated by high-level planner\\
    \bottomrule
    \end{tabular}
    \vspace{5pt}
    \caption{Observation space of our framework in the real-world.}
    \label{app:observation_real}
\end{table}

\subsection{Resets}
In the real-world evaluation, we used a trajectory from the ARCTIC dataset as our goal trajectory. In terms of resets in our real-world experiments, we pose the object within 3 centimeters and 0.5 radians of the initial pose of the goal trajectory during resets. We evaluated 20 times and reported the completion rate.
\subsection{Evaluation}
When evaluating real-world experiments, we take the same metric (completion rate) as in simulation. It will fail if the tolerance is exceeded($5$ centimeters in translation, $2.5$ centimeters in object's longest dimension multiplied by rotation angle, and $0.5$ radians in joint angle), and record the completion rate.

\section{Hyperparameters of the PPO}
Table.\ref{app:hyper_ppo} gives the hyperparameters of the PPO.

\begin{table}[thbp]
    \centering
    \begin{tabular}{c|c}
    \toprule
    \rowcolor[HTML]{D4F7EE}
    Hyperparameters & Value  \\\hline
    Num mini-batches & 4 \\
    Num opt-epochs & 5 \\
    Num episode-length & 8\\\hline
    Hidden size & [1024, 1024, 512, 256] \\
    Clip range & 0.2\\
    Max grad norm & 1\\
    Learning rate & 3.e-4 \\
    Discount ($\gamma$) & 0.998\\
    GAE lambda ($\lambda$) & 0.95\\
    Init noise std & 0.8\\\hline
    Desired kl & 0.02 \\
    Ent-coef & 0\\
    \bottomrule
    \end{tabular}
    \vspace{5pt}
    \caption{Hyperparameters of PPO.}
    \label{app:hyper_ppo}
\end{table}

\section{Domain Randomization}

Isaac Gym provides lots of domain randomization functions for RL training. We add the randomization for all the tasks as shown in Table.~\ref{tab:dr} for each environment. we generate new randomization every 1000 simulation steps.

\begin{table}[thbp]
\centering
            \begin{tabular}{c|c|c|c}
            \toprule
            \rowcolor[HTML]{D4F7EE}
            Parameter & Type & Distribution & Initial Range\\
            \midrule
            
            \rowcolor[HTML]{EFEFEF} 

            \multicolumn{4}{l}{\textbf{Robot}} \\
            Mass & Scaling & $ \textrm{uniform}$ & [0.5, 1.5] \\ %
            Friction & Scaling & $ \textrm{uniform}$ & [0.7, 1.3] \\
           Joint Lower Limit & Scaling & $ \textrm{loguniform}$ & [0.0, 0.01] \\ %
            Joint Upper Limit & Scaling & $ \textrm{loguniform}$ & [0.0, 0.01] \\
            Joint Stiffness & Scaling & $ \textrm{loguniform}$ & [0.0, 0.01] \\
            Joint Damping & Scaling & $ \textrm{loguniform}$ & [0.0, 0.01] \\

            \rowcolor[HTML]{EFEFEF} 
            \multicolumn{4}{l}{\textbf{Object}} \\
            Mass & Scaling & $ \textrm{uniform}$ & [0.5, 1.5]  \\
            Friction & Scaling & $ \textrm{uniform}$ & [0.5, 1.5] \\
            Scale & Scaling & $ \textrm{uniform}$ & [0.95, 1.05] \\
            Position Noise & Additive & $ \textrm{gaussian}$ & [0.0, 0.02] \\
            Rotation Noise & Additive & $ \textrm{gaussian}$ & [0.0, 0.2] \\
            
            \rowcolor[HTML]{EFEFEF} 
            \multicolumn{4}{l}{\textbf{Observation}} \\
            Obs Correlated. Noise & Additive & $ \textrm{gaussian}$ & [0.0, 0.001] \\
            Obs Uncorrelated. Noise & Additive & $ \textrm{gaussian}$ & [0.0, 0.002] \\

            \rowcolor[HTML]{EFEFEF} 
            \multicolumn{4}{l}{\textbf{Action}} \\
            Action Correlated Noise & Additive & $ \textrm{gaussian}$ & [0.0, 0.015]  \\
            Action Uncorrelated Noise & Additive & $ \textrm{gaussian}$ & [0.0, 0.05] \\

            \rowcolor[HTML]{EFEFEF} 
            \multicolumn{4}{l}{\textbf{Environment}} \\
            Gravity & Additive & $ \textrm{normal}$ & [0, 0.4] \\

            \bottomrule
            \end{tabular}
    \vspace{5pt}
    \caption{Domain randomization of all the tasks.}
     \label{tab:dr}
\end{table}

\section{Goal Representation}
\begin{figure*}[!t]
    \centering
    \includegraphics[width=0.9\linewidth]{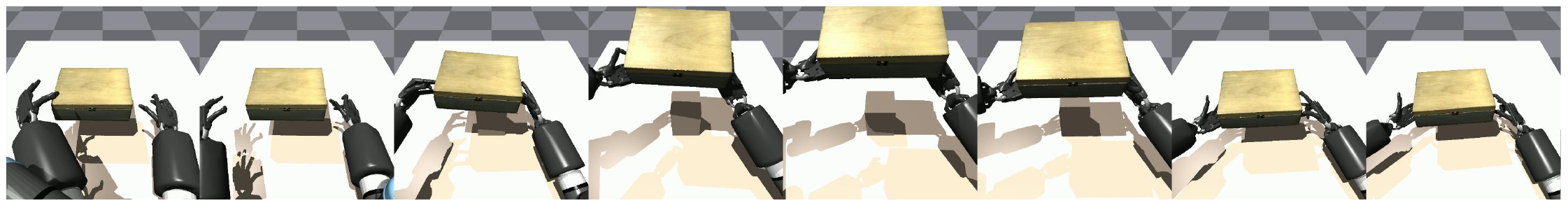}
    \caption{Snapshots of the goal representation experiment.}
    \label{fig:snapshot_rebuttal}
\end{figure*}

Our framework requires the full 6D pose of the object trajectory at each time-step. For downstream tasks, we are able to get the trajectories in many ways. For example, we can specify a few key poses of the object based on the task, and use linear interpolation to generate the pose trajectories between the key poses. Another solution is to use some object motion synthesis methods in the field of graphics to generate the 6D pose of the object trajectory, such as~\cite{li2023controllable}.
We design an additional experiment to prove that our method can work with this setup. Taking a task “lifting up the box in the air and dropping it down” as an example, we can just simply manually define the pose of the box in the air and the landing point after picking it up as the key poses, then interpolate it as our goal trajectory, and train the robot to manipulate the object to follow the goal trajectory. The results are shown in Table~\ref{app:goal_rep} and the snapshot are shown in Figure~\ref{fig:snapshot_rebuttal}. The results show that we can complete the task under this setting, and shows that our framework is extensible.
\begin{table}[thbp]
\centering
\begin{tabular}{l|cc}
\toprule
    &  Box & Manually Designed Trajectory \\ \midrule
Our & \pms{100}{0.0} &  \pms{100}{0.0}        \\ \bottomrule
\end{tabular}
\vspace{5pt}
\caption{Results for the goal representation experiments.}
\label{app:goal_rep}
\end{table}

\section{Time-indexed of the Goal Trajectory}
For the time-indexed of the goal trajectory, we randomize the sampling of input trajectories with various time gaps during training. We added an additional experiment to show that varying the time gap does not significantly affect our performance, indicating that our approach maintains generalization despite these temporal constraints. Using Coffee Maker as an example, we interpolated the goal trajectories provided by ARCTIC datasets to varying levels, and tested the completion rate of our policy on it. Origin represents the original goal trajectory from ARCTIC, and Skip 2 and Skip 1 represent skipping 2/1 poses between every two poses on the basis of origin goal trajectory. Inter 2 and Inter 1 represent insertion of 2/1 pose between every two poses according to the linear interpolation method on the basis of origin goal trajectory. Our policy generates robot actions conditioned on a sequence of object goal poses, with the time gaps defined by the distance between each consecutive goal pose. In this experiment, we randomly use time gaps of 0 to 3 units when sampling the object trajectory to test whether the policy is sensitive to these time gaps (Random Sample). The results is shown in Table~\ref{app:time_index}, the performance of the trained policy will not vary greatly.
\begin{table}[thbp]
\centering
\begin{tabular}{l|cccccc}
\toprule
    &  Skip 2             &  Skip 1             &  Origin        &  Inter 1            &  Inter 2 & Random Sample \\ \midrule
Our & \pms{85.7}{7.5} & \pms{86.5}{3.4} & \pms{86.1}{5.5} & \pms{85.4}{8.1} & \pms{87.2}{4.8}           & \pms{86.2}{6.2}          \\ \bottomrule
\end{tabular}
\vspace{5pt}
\caption{Results for the time-indexed experiments.}
\label{app:time_index}
\end{table}

\section{Model-base Baselines}
We add a baseline of the model-based method in the simulation, the result is shown in Table~\ref{app:mpc}. We use the sample-based model predictive control method that is modified from~\cite{pezzato2023samplingbased} as our low-level controller in our tasks. Our method outperforms the sample-based model predictive control by 46.3\% on average.
\begin{table}[!t]
\centering
% \scriptsize
\begin{tabular}{l|ccccccccc}
\toprule
 % &
 %  Box &
 %  Capsule &
 %  Espresso &
 %  Ketch &
 %  Micro &
 %  Mixer &
 %  Note &
 %  Scis &
 %  % Waffle &
 %  Laptop \\
 % &
 %   &
 %  Machine &
 %  Machine &
 %  up &
 %  wave &
 %   &
 %  book &
 %  sors &
 %  % iron &
  
& \multirow{2}{*}{Box} & \multirow{2}{*}{\shortstack{Coffee \\ Maker}} & \multirow{2}{*}{\shortstack{Espresso \\Machine}} & \multirow{2}{*}{Ketch} & \multirow{2}{*}{\shortstack{Micro- \\ wave}} & \multirow{2}{*}{Mixer} & \multirow{2}{*}{\shortstack{Note- \\ book}} & \multirow{2}{*}{\shortstack{Scis- \\ sors}} & \multirow{2}{*}{Laptop}

   \\ \\
  
\midrule
MPC &

  \multicolumn{1}{r}{\pms{31.6}{1.5}} &
  \multicolumn{1}{r}{\pms{22.8}{0.7}} &
  \multicolumn{1}{r}{\pms{30.4}{2.3}} &
  \multicolumn{1}{r}{\pms{24.8}{1.3}} &
  \multicolumn{1}{r}{\pms{28.6}{1.9}} &
  \multicolumn{1}{r}{\pms{26.8}{0.5}} &
  \multicolumn{1}{r}{\pms{19.2}{2.0}} &
  \multicolumn{1}{r}{\pms{23.6}{0.8}} &
  \multicolumn{1}{r}{\pms{24.0}{1.4}} \\
Ours &
  \multicolumn{1}{r}{\pms{\textbf{100}}{0.0}} &
  \multicolumn{1}{r}{\pms{\textbf{86.1}}{5.5}} &
  \multicolumn{1}{r}{\pms{\textbf{81.1}}{8.6}} &
  \multicolumn{1}{r}{\pms{\textbf{41.2}}{13.3}} &
  \multicolumn{1}{r}{\pms{\textbf{100}}{0.0}} &
  \multicolumn{1}{r}{\pms{\textbf{57.6}}{4.9}} &
  \multicolumn{1}{r}{\pms{\textbf{38.7}}{3.3}} &
  \multicolumn{1}{r}{\pms{\textbf{41.4}}{14.9}} &
  \multicolumn{1}{r}{\pms{\textbf{100}}{0.0}} \\
  \bottomrule
\end{tabular}
\vspace{5pt}
\caption{Results for the experiments of using one policy per object.}
\label{app:mpc}

\vspace{-0.5cm}
\end{table}

\end{document}